\documentclass[11pt]{article}
\usepackage[letterpaper,hmargin=.85in,vmargin=.80in]{geometry}
\usepackage[authoryear,round]{natbib}

\usepackage{times}
\usepackage[T1]{fontenc}
\usepackage[utf8]{inputenc}
\usepackage{amsmath,amssymb,amsthm,mathtools}
\usepackage{microtype}
\usepackage{graphicx}
\usepackage{subcaption}
\usepackage{booktabs,array,tabularx}
\usepackage{float}
\floatstyle{ruled}
\newfloat{algorithm}{tbp}{loa}
\floatname{algorithm}{Algorithm}
\usepackage{enumitem}
\usepackage{etoc}
\usepackage{xcolor}
\newsavebox{\resultbox}
\usepackage{tikz}
\usetikzlibrary{arrows.meta,positioning,calc}
\usepackage{url}
\usepackage{multirow}
\usepackage{hyperref}
\usepackage{comment}
\hypersetup{
    colorlinks=true,
    linkcolor=magenta,    
    citecolor=blue,    
    urlcolor= teal     
}
\hypersetup{pdftitle={Amortizing Tempered Sampling with Corrected Self-Training},
  pdfauthor={Hoang-Viet Nguyen, Hai-Dang Dau, Alexandre H. Thiery},pdfsubject={Corrected self-training for tempered sampling}}
\newcommand{\method}{\textsc{SSS}}
\newcommand{\TABG}{TA-BG}
\newcommand{\PITA}{PITA}

\newcommand{\E}{\mathbb E}

\newcommand{\Normal}{\mathcal N}

\newcommand{\Data}{\mathcal D}
\newcommand{\dd}{\mathrm d}
\newcommand{\lse}{\operatorname{logsumexp}}
\newcommand{\norm}[1]{\left\lVert #1\right\rVert}

\newcommand{\NA}{\textemdash}
\renewcommand{\epsilon}{\varepsilon}

\theoremstyle{remark}

\title{Amortizing Tempered Sampling with\\Corrected Self-Training}
\author{Hoang-Viet Nguyen$^{1}$ \qquad Hai-Dang Dau$^{2}$ \qquad Alexandre H. Thiéry$^{1}$\\[4pt]
\small $^{1}$National University of Singapore \qquad $^{2}$Nanyang Technological University}
\date{}

\begin{document}

\maketitle

\begin{abstract}
Learning a neural sampler can reduce the cost of repeated sampling from an un-normalized distribution, but obtaining good training data remains difficult. We develop a {\it corrected self-training} scheme across a sequence of tempered targets, making use of importance sampling and Markov chain Monte Carlo to correct the samples and replay buffers to store them. We introduce Sequential Score Sampler (\method{}) as a particular case. In \method{}, the proposal is a score-based diffusion trained by regression on energy gradients; and the samples are corrected by importance weights on the path space.
After training, samples can be collected directly from the model, from its importance-corrected proposals, or from the final-temperature buffer. The method amortizes repeated sampling by encoding the global structure of the target in the model. SSS achieves competitive sampling accuracy across the benchmarks and, on several targets, uses shorter runtimes with fewer energy evaluations than competing samplers.
\end{abstract}

\section{Introduction}
\label{sec:intro}

Many inference and simulation problems require repeated sampling from a distribution whose density is known only up to a normalizing constant. %
Multimodality makes this difficult: local Markov chain Monte Carlo (MCMC) %
may explore a basin thoroughly while rarely moving between basins \citep{woodard2009conditions}. Annealed importance sampling and sequential Monte Carlo (SMC) address this difficulty through intermediate distributions \citep{neal2001annealed,chopin2002sequential,delmoral2006smc,dai2022invitation}. However, obtaining further accurate estimates still requires additional sampling computation. Can some of that computation instead be invested in learning a reusable proposal?

Learned samplers pursue this amortization \citep{zhang2022path,vargas2023denoising,midgley2023flow,akhoundsadegh2024iterated}. Once trained, a generative model
can generate new proposal samples without continuing a long local Markov chain. The difficulty is how to train it without representative target samples. One approach is to generate samples from the current model, correct for proposal mismatch, and use the resulting data to update it.
Importance weighting supplies this correction by adjusting the samples' contributions without requiring the target's normalizing constant.

We develop corrected self-training along a sequence of tempered targets and instantiate it with Sequential Score Sampler (\method{}). Rather than fit a separate proposal at each temperature, we train one model that takes the temperature as an input. Initial sample populations can come from a suitable sampler or existing problem-specific simulations. During training, the model generates proposals at each intermediate temperature. We resample them according to their importance weights and apply MCMC steps that preserve the corresponding target. Reweighting to the next temperature then provides data for its buffer. These buffers retain weighted samples from recent iterations, allowing to reuse costly sampling results over multiple optimization steps instead of training only on the latest batch. Each outer iteration updates the model using data from all temperatures.

In SMC with local Markov transitions, reweighting and resampling cannot create configurations outside the current population; reaching an unrepresented basin therefore depends on rare transitions. A learned global proposal can instead reuse information encoded from earlier samples and across temperatures \citep{dibak2022temperature}, without starting from a surviving particle. Problem-specific architecture can support this reuse. For interacting-particle systems, for example, an equivariant network shares learned geometric relations across rotations and permutations of identical particles \citep{satorras2021en}. Such structure reduces the need to learn equivalent configurations separately and can support proposals beyond those stored in a finite population.

Section~\ref{sec:method} presents the general training scheme, independently of the proposal family and initialization sampler. Section~\ref{sec:score} gives the score-based implementation. {Section~\ref{sec:experiments} asks how much accuracy is retained by the learned proposal and how much is gained by correcting its samples or drawing from the final buffer. We compare all three outputs with Temperature-Annealed Boltzmann Generators (\TABG{}; \citealp{schopmans2025temperature}), Progressive Inference-Time Annealing (\PITA{}; \citealp{akhoundsadegh2025progressive}), and Adjoint Sampling (AS; \citealp{havens2025adjoint}).}

\section{Corrected self-training across temperatures}
\label{sec:method}

\subsection{Targets and initialization}
\label{sec:targets}

Let $\rho$ be a normalized reference density from which sampling is easy and whose support covers the target. We write the target as $\pi(x)=Z_1^{-1}\rho(x)e^{-E(x)}$ and define the intermediate unnormalized and normalized densities
\begin{equation}
  \bar\pi_\beta(x)=\rho(x)e^{-\beta E(x)},
  \qquad \pi_\beta(x)=\frac{\bar\pi_\beta(x)}{Z_\beta},
  \qquad 0\leq\beta\leq1.
  \label{eq:targets}
\end{equation}
Here $Z_\beta=\int\bar\pi_\beta(x)\,\dd x$, so $\pi_0=\rho$ and $\pi_1=\pi$.
We call $\beta$ the `temperature' parameter. Choose a ladder $0=\beta_0<\cdots<\beta_K=1$ and obtain initial weighted populations approximating the targets at its levels. These choices may use problem knowledge, existing simulation data, or a preliminary sampler such as SMC or parallel tempering. Adjacent targets should overlap sufficiently for incremental importance reweighting. Our adaptive-SMC initialization is one particular implementation of this step.%

Let $Q_\theta^\beta$ denote the distribution generated by the learned proposal when given temperature $\beta$. The same parameters $\theta$ are shared across all levels; in a neural sampler, they are the network weights. We write $q_\theta(x\mid\beta)$ for its density.

For each target $\pi_{\beta_k}$, we maintain a weighted training dataset $\Data_k$, called a replay buffer, initially filled with the corresponding population. Retaining data across refreshes is useful when generating corrected samples is expensive relative to a model update: each sample can be reused in several optimization steps, and training need not rely on a single, potentially noisy batch. Pooling recent batches also retains configurations from earlier iterations while the proposal changes.

\begin{figure}[tbp]
\centering
\begin{tikzpicture}[
  font=\small,
  box/.style={draw,rounded corners=2pt,align=center,inner sep=6pt},
  arr/.style={-{Stealth[length=2mm]},semithick}]
  \node[box,text width=2.45cm] (data) at (-5.1,0)
    {weighted training buffers\\$\Data_0,\ldots,\Data_K$};
  \node[box,fill=black!5,text width=2.4cm] (model) at (-1.1,0)
    {One model for all temperatures\\$Q_\theta^\beta$};
  \node[box,text width=4.65cm] (correct) at (4.60,0)
    {Importance correction at $\beta_k$;\\MCMC exploration at $\beta_k$;\\reweight towards $\beta_{k+1}$};
  \draw[arr] (data) -- node[above,font=\footnotesize,align=center]{train\\jointly} (model);
  \draw[arr] (model) -- node[above,font=\footnotesize,align=center]{fresh\\proposals} (correct);
  \coordinate (turn) at ($(correct.south)+(0,-0.65)$);
  \draw[arr] (correct.south) -- (turn) -- node[midway,below=2pt,font=\footnotesize]{refresh $\Data_{k+1}$} (data.south |- turn) -- (data.south);
\end{tikzpicture}
\caption{Corrected self-training across temperatures.} %
\label{fig:overview}
\end{figure}
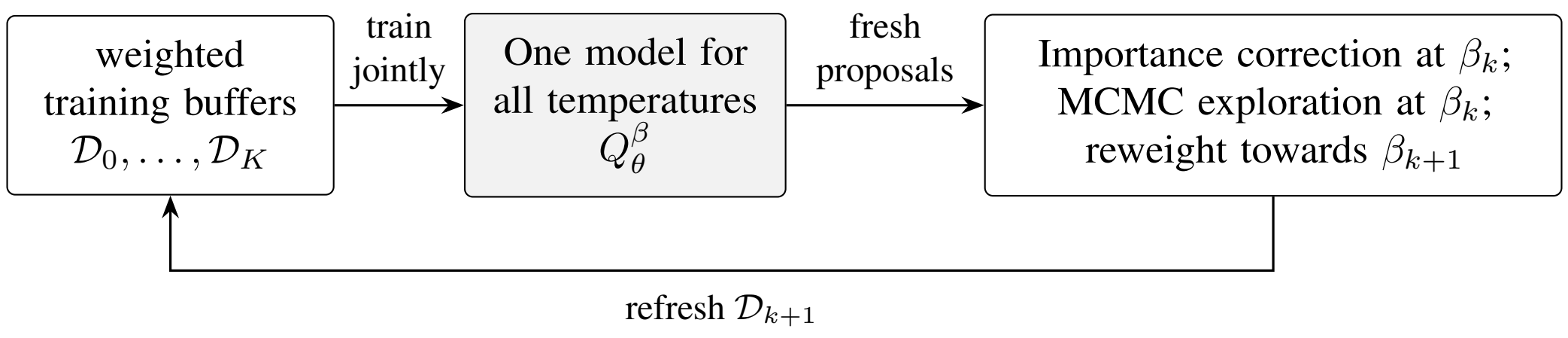

Training alternates between refreshing these datasets with corrected proposals and fitting the shared model to data from all temperatures. A refresh at $\beta_k$ supplies data for $\Data_{k+1}$. Figure~\ref{fig:overview} summarizes the data flow, and Algorithm~\ref{alg:train} specifies the update order.

\subsection{Generating corrected training data}
\label{sec:refresh}
For each selected level $k\in\{1,\ldots,K-1\}$, we refresh $\Data_{k+1}$ using corrected proposals from the preceding, typically easier target $\pi_{\beta_k}$. Well-initialized buffers may be kept fixed; our choice of refreshed levels is specified in Section~\ref{sec:exp-setup}.

\subsubsection{Generating weighted proposals at \texorpdfstring{$\beta_k$}{beta k}}
\label{subsect:local}

At fixed model parameters $\theta$, generate $N$ proposal samples $x_i$ and nonnegative importance weights $w_i$ targeting $\pi_{\beta_k}$. We require the importance-sampling consistency property: for test functions $f$,
\begin{equation}
  \widehat\mu_{\beta_k,N}(f)
  =\frac{\sum_{i=1}^{N}w_i f(x_i)}{\sum_{i=1}^{N}w_i}
  \longrightarrow \E_{\pi_{\beta_k}}[f]
  \qquad\text{as }N\to\infty.
  \label{eq:weighted-estimate}
\end{equation}
When the proposal density is available, the weight is simply
$w_i=\bar\pi_{\beta_k}(x_i)/q_\theta(x_i\mid\beta_k)$.
Otherwise, importance sampling can be performed on an extended space that includes the sampler's auxiliary variables, provided the extended target has marginal $\pi_{\beta_k}$. Section~\ref{sec:sampler} gives such a construction for diffusion paths.

\subsubsection{Resampling and applying MCMC}
\label{subsect:mcmc}
We resample the proposals according to their importance weights and apply $\pi_{\beta_k}$-invariant MCMC updates. The weights adjust the proposals' contributions to the target approximation. The Markov moves can diversify duplicated particles and explore nearby configurations while preserving the target distribution.%

\subsubsection{Reweighting and updating the buffer at \texorpdfstring{$\beta_{k+1}$}{beta k+1}}
\label{subsect:buffer}
To turn the resulting population into training data for the next target, assign each sample the incremental importance weight
\begin{equation}
  v_{k\to k+1}(x)
  =\frac{\bar\pi_{\beta_{k+1}}(x)}{\bar\pi_{\beta_k}(x)}
  =\exp\!\Big\{-(\beta_{k+1}-\beta_k) \, E(x)\Big\}.
  \label{eq:bridge}
\end{equation}
Unlike the proposal-correction weights $w_i$, these weights account only for the change of target from $\beta_k$ to $\beta_{k+1}$. We insert this weighted batch into $\Data_{k+1}$. In our implementation, each buffer has fixed capacity and uses first-in, first-out (FIFO) replacement: new batches gradually replace the oldest entries. This retains data across several refreshes without keeping all past populations indefinitely. Before insertion, we normalize weights within each incoming batch to have mean one, giving equal-sized batches equal total weight. Appendix~\ref{app:replay} gives the details.

\subsection{Learning one model from all temperatures}
\label{sec:joint}

After refreshing the selected buffers, we draw training samples from every $\Data_k$ with probabilities proportional to their stored weights. Each temperature contributes the same number of samples to a training minibatch. The fitting objective depends on the proposal family: a normalizing flow can be trained by weighted maximum likelihood, whereas \method{} uses the score-regression loss in Section~\ref{sec:regression}. Gradients are taken only through this fitting objective, not through sample generation, importance weights, resampling, or MCMC. Sharing parameters allows features learned at one temperature to influence proposals at others, while retaining training data from the entire ladder throughout optimization.

\begin{algorithm}[t]
\caption{Corrected self-training across temperatures}
\label{alg:train}
\begin{enumerate}[label=\arabic*:,nosep,leftmargin=2em]
\item[] \textbf{Input:} Reference $\rho$, energy $E$, ladder $\{\beta_k\}_{k=0}^{K}$, initialized weighted buffers $\{\Data_k\}_{k=0}^{K}$ %
\item Initialize the shared model parameters $\theta$.
\item \textbf{for} each outer iteration \textbf{do}
\item \hspace*{1em}\textbf{for} each selected level $k$ \textbf{in parallel} \textbf{do}
\item \hspace*{2em}Generate $N$ proposals and importance weights at $\beta_k$ using $Q_\theta^{\beta_k}$
\item \hspace*{2em}Resample by these weights; apply $\pi_{\beta_k}$-invariant MCMC updates.
\item \hspace*{2em}Compute $v_i=\bar\pi_{\beta_{k+1}}(x_i)/\bar\pi_{\beta_k}(x_i)$.
\item \hspace*{2em}Update $\Data_{k+1}$ with this weighted batch (Section~\ref{subsect:buffer}).
\item \hspace*{1em}Update $\theta$ using weighted minibatches from all buffers.
\item \textbf{return} The trained proposal, its importance-weight calculation, and $\Data_K$.
\end{enumerate}
\end{algorithm}

\subsection{Final sample collection}
\label{sec:final-samples}

After training, the construction supports three ways to collect samples at the target temperature, evaluated in Section~\ref{sec:experiments} as \method{} (raw), \method{} (corrected), and \method{} (buffer).

\textbf{Raw model samples.}
Draw directly from $Q_\theta^1$, without importance correction or MCMC. The sampling cost is that of the proposal itself, but its distribution may differ from the target.

\textbf{Importance-corrected samples.}
Generate proposals at $\beta=1$ together with their importance weights. These can be used in weighted estimates or to resample the proposals; we evaluate the latter option. The weighted estimates satisfy~\eqref{eq:weighted-estimate}, but poor proposal-target overlap can concentrate the weights on a few samples, leading to low effective sample size and many duplicates after resampling.

\textbf{Final-buffer samples.}
Draw from $\Data_K$ according to its stored weights. This population has already undergone the correction, MCMC exploration, and temperature-bridging steps during training. Its pool of distinct configurations is limited by the buffer capacity; obtaining new configurations requires further buffer updates.

\section{A score-based implementation}
\label{sec:score}

{
Sequential Score Sampler (\method{}) implements the proposal
$Q_\theta^\beta$ of Section~\ref{sec:method} using a
score-based diffusion. A single network $r_\theta(z,t,\beta)$
is shared across temperatures, where $\beta$ specifies the
tempered target and $t$ the diffusion noise level.
We describe how the network is trained and how the resulting
diffusion generates proposals with computable importance weights.
}

\subsection{Learning the relative score}
\label{sec:regression}

{
For a given temperature $\beta$, let $X\sim\pi_\beta$.
Let $\pi_{\beta,t}$ denote the density of the Gaussian pertubation $X_t=\alpha_t X+\sigma_t\epsilon$ for $\alpha_t=e^{-t}$ and $\sigma_t^2=1-\alpha_t^2$ and $\epsilon\sim\Normal(0,I_d)$ where $t>0$ controls the noise level. We train the network to predict the score of this noisy distribution relative to the standard Gaussian distribution $\rho$:
\begin{equation}
\label{eq:relative-score}
r^*(z,t,\beta)
=\nabla_z \log [\pi_{\beta,t}(z) / \rho(z)]
=\nabla_z \log \pi_{\beta,t}(z)+z.
\end{equation}
Thus, $r^*$ is the correction to the known Gaussian score and it can be learned with a standard regression loss. Two standard choices of regression target are:
\begin{equation}
\varphi_{\rm DSM} =X_t-\epsilon/\sigma_t,
\qquad \textrm{and} \qquad
\varphi_{\rm TSM} =\alpha_t\bigl(\nabla\log\pi_\beta(X)+X\bigr).
\end{equation}
Denoising score matching (DSM) uses the noise added to $X$
\citep{vincent2011connection,song2021score}, while
target score matching (TSM) uses the target score at $X$
\citep{debortoli2024target}.
Both have conditional expectation $r^*(z,t,\beta)$
given $X_t=z$, so either can be used to learn the
same function.
For a standard Gaussian reference $\rho$,
the TSM target simplifies to
$\varphi_{\rm TSM}=-\alpha_t\beta\nabla E(X)$
by~\eqref{eq:targets} (Appendix~\ref{app:theory}).
The energy gradient can be stored with $X$ and reused
at different noise levels.

In practice, we train on the weighted replay buffers
$\Data_0,\ldots,\Data_K$ rather than exact target samples.
For each training example, we select $k$ uniformly,
draw $X$ from $\Data_k$ according to its stored weight,
and sample $t$ and $\epsilon$ to construct $X_t$ and
the regression target $\varphi$.
The training loss is
\begin{equation}
\mathcal{L}(\theta)
=\E_{k,t,X,\epsilon}\!\left[
    \lambda(t,\beta_k)
    \norm{r_\theta(X_t,t,\beta_k)-\varphi}^2
  \right],
\label{eq:loss}
\end{equation}
where $\lambda(t,\beta_k)>0$ controls the contribution
of each noise level and temperature.

With exact target samples and sufficient model capacity,
either regression target recovers $r^*$ at the population
level. In practice, errors in the buffers' coverage or
mode weights can affect the learned score.
Section~\ref{sec:exp-setup} specifies the loss used
in each experiment.}

\subsection{Generating samples and computing weights}
\label{sec:sampler}
{We first give the sampler and its weights for a standard Gaussian tempering reference $\rho$. In this case,} the learned relative score defines the reverse-time diffusion
\begin{equation}
  Y_0\sim\rho,\qquad
  \dd Y_s=\bigl[-Y_s+2r_\theta(Y_s,T-s,\beta)\bigr]\,\dd s
           +\sqrt{2}\,\dd W_s,
  \quad 0\leq s\leq T.
  \label{eq:backward_sde}
\end{equation}
With an accurate relative score and a horizon $T$ for which $\pi_{\beta,T}$ is close to $\rho$, its terminal distribution approximates $\pi_\beta$. We use an exponential integrator that treats the Gaussian-reference dynamics exactly. Appendix~\ref{app:numerics} gives the numerical update and its implementation.
The numerical sampler produces a path $y_{0:L}$ starting at $y_0\sim\rho$, with endpoint $x=y_L$. For $h=T/L$ and $a=e^{-h}$, its Gaussian transition density is $q_{\theta,n}(\cdot\mid y,\beta) =\Normal\!\left(ay+2(1-a)r_\theta(y,T-nh,\beta),\,(1-a^2)I_d\right)$.
Let $g_h(y'\mid y)=\Normal(y';ay,(1-a^2)I_d)$ be the exact Ornstein--Uhlenbeck reference transition. A chain started at $\rho$ and propagated with $g_h$ still has endpoint distribution $\rho$. Multiplying its path density by $e^{-\beta E(y_L)}$ therefore gives an extended target with endpoint marginal $\bar\pi_\beta$. The ratio of this unnormalized path density to the proposal path density is
\begin{equation}
  w(y_{0:L})
  =e^{-\beta E(y_L)}
    \prod_{n=0}^{L-1}
    \frac{g_h(y_{n+1}\mid y_n)}
         {q_{\theta,n}(y_{n+1}\mid y_n,\beta)}.
  \label{eq:path-weight}
\end{equation}
For fixed $\theta$ and integrable $f$, this construction gives
$\E[w(y_{0:L})f(y_L)]=Z_\beta\E_{\pi_\beta}[f]$, where the first expectation is over proposal paths. Independent simulations therefore satisfy~\eqref{eq:weighted-estimate}. The correction applies to the discrete sampler actually used: it requires neither an exact score nor evaluation of the endpoint density $q_\theta(x\mid\beta)$. The weights can be accumulated online without storing the path (Appendix~\ref{app:numerics}).

\section{Experiments}
\label{sec:experiments}

\subsection{Targets, network architectures, and baselines}
\label{sec:exp-setup}

\paragraph{Targets.}
{We study five targets. Two factorized Gaussian mixtures, GMM16 and GMM50, have the same one-dimensional components but differ in dimension: the number of modes grows from $2^{16}$ to $2^{50}$. Both admit exact reference draws. DW-4 and LJ-13 test whether the sampler reproduces arrangements of interacting particles \citep{kohler2020equivariant}. DW-4 has several preferred configurations; LJ-13 also has steep repulsion when particles come close. Finally, we study Alanine Dipeptide at 300 K, where the backbone angles distinguish molecular conformations \citep{akhoundsadegh2025progressive}. Appendix~\ref{app:target-definitions} gives the target definitions and reference samples.}

\paragraph{Initialization and rejuvenation.}
In all experiments, one adaptive SMC run selects the temperature ladder and initializes the buffers \citep{delmoral2006smc,jasra2011adaptive,beskos2016convergence}. We use {Hamiltonian Monte Carlo (HMC), with an empirical preconditioner for the mixtures, DW-4 and LJ-13 and an identity preconditioner in internal coordinates for Alanine Dipeptide,} for local rejuvenation; its tuning and the reuse of the SMC-tuned parameters are specified in Appendix~\ref{app:smc}. The reference buffer $\Data_0$ contains independent $\rho$-samples. The first bridge weights such samples by $e^{-\beta_1E(x)}$, followed by SMC resampling and Markov moves, to initialize $\Data_1$. We keep $\Data_0$ and $\Data_1$ fixed: the former already contains exact reference draws, and the latter provides an SMC-based training population at the first nontrivial level without relying on a learned proposal. Only $\Data_2,\ldots,\Data_K$ are refreshed, using levels $k=1,\ldots,K-1$ in Algorithm~\ref{alg:train}.

\paragraph{Network architecture.}
In all experiments, a single network $r_\theta(x_t, t, \beta)$ is shared across all temperatures of the ladder, modeling the relative score $r^*(x_t, t, \beta)$ described in Section~\ref{sec:regression}. We use a residual multilayer perceptron (MLP) for the Gaussian mixtures, and an E($n$)-equivariant graph neural network (EGNN)~\citep{satorras2021en} for DW-4 and LJ-13. {For Alanine Dipeptide we use a diffusion transformer with one token per atom.} Appendix~\ref{app:architecture} gives the details.

\paragraph{{Training loss.}}
{We use target score matching on the Gaussian mixtures, DW-4, and LJ-13. For Alanine Dipeptide we use denoising score matching across the full range of diffusion times $[t_{\min},T]$. At low noise ($t<t_{\mathrm{threshold}}$), we add a small target score matching term so that the network also learns from the energy gradient of the clean configuration. Appendix~\ref{app:aldp-loss} gives the loss and its settings.}

\paragraph{Baselines.}
{We compare \method{} with three methods that learn to sample from an energy function. \TABG{} and \PITA{} share our use of a temperature ladder. \TABG{} first fits a normalizing flow at high temperature. It then reweights its samples to the next temperature and fits the flow to the resampled population \citep{schopmans2025temperature}. \PITA{} trains a diffusion at each temperature and uses an auxiliary energy model to compute weights along the reverse diffusion, producing training samples for the next temperature \citep{akhoundsadegh2025progressive}.

AS shares a different ingredient with \method{}: it reuses endpoint gradients to train a diffusion \citep{havens2025adjoint}. AS stores endpoints generated by its current sampler. For each endpoint it can draw fresh intermediate states from the base diffusion bridge and regress the control against the stored terminal-cost gradient. In \method{}, the endpoints used for training come from buffers at several temperatures. The diffusion supplies its own path weights through~\eqref{eq:path-weight}, and one network learns from the corrected populations across the ladder.

For the two temperature-based baselines, we match the ladders, account for network size, and use the same nominal HMC proposal budget on GMM16, GMM50, and DW-4 (Appendix~\ref{app:hyperparameters}). For AS we use an HMC-augmented reimplementation on all five targets, described in the same Appendix section.}

\paragraph{Cost and evaluation.}
{We compare each sample set with reference draws using summaries appropriate to the target. For the mixtures, we compare pooled coordinates and energies. For DW-4 and LJ-13, we compare pairwise distances, energies, and whole configurations after alignment. For Alanine Dipeptide, we compare backbone angles and energies, following \citet{akhoundsadegh2025progressive}. Appendix~\ref{sec:reported-metrics} defines each discrepancy. We report energy and gradient evaluations alongside timing measurements in Appendix~\ref{app:costs}.}

\subsection{Main results}
\begin{table}[!t]
\centering\small
\caption{{Sampler performance (mean $\pm$ standard deviation; lower is better). \textbf{Bold} marks the lowest displayed mean and \underline{underlining} the second-lowest distinct mean. Equal displayed means share a rank. A \NA{} denotes an inapplicable metric and $\ast$ a mean above $10^4$. LJ-13 results for \TABG{} and \PITA{} are from \citet{akhoundsadegh2025progressive} and retain the published precision.}}
\label{tab:reported}
\setlength{\tabcolsep}{5pt}
\begin{lrbox}{\resultbox}\begin{tabular}{@{}lrrrr@{}}
\toprule
Target & Method & Distance-$\mathcal W_2\downarrow$ & Energy-$\mathcal W_2\downarrow$ & Geometric-$\mathcal W_2\downarrow$ \\
\midrule
GMM16 ($d=16$) & TA-BG & $\underline{0.023 \pm 0.005}$ & $0.067 \pm 0.043$ & \NA{} \\
 & PITA & $0.168 \pm 0.030$ & $0.984 \pm 0.082$ & \NA{} \\
 & AS & $2.221 \pm 0.072$ & $9.947 \pm 0.379$ & \NA{} \\
 & SSS (raw) & $0.130 \pm 0.091$ & $1.153 \pm 0.121$ & \NA{} \\
 & SSS (corrected) & $\mathbf{0.005 \pm 0.005}$ & $\mathbf{0.040 \pm 0.004}$ & \NA{} \\
 & SSS (buffer) & $0.041 \pm 0.009$ & $\underline{0.048 \pm 0.016}$ & \NA{} \\
\midrule
GMM50 ($d=50$) & TA-BG & $\underline{0.114 \pm 0.153}$ & $1.065 \pm 1.640$ & \NA{} \\
 & PITA & $2.115 \pm 0.021$ & $240.014 \pm 1.016$ & \NA{} \\
 & AS & $2.338 \pm 0.060$ & $32.965 \pm 1.294$ & \NA{} \\
 & SSS (raw) & $0.166 \pm 0.034$ & $4.304 \pm 0.807$ & \NA{} \\
 & SSS (corrected) & $\mathbf{0.048 \pm 0.022}$ & $\mathbf{0.321 \pm 0.059}$ & \NA{} \\
 & SSS (buffer) & $0.208 \pm 0.037$ & $\underline{1.025 \pm 0.201}$ & \NA{} \\
\midrule
DW-4 ($d=8$) & TA-BG & $0.317 \pm 0.016$ & $24.519 \pm 10.463$ & $0.789 \pm 0.022$ \\
 & PITA & $\mathbf{0.150 \pm 0.124}$ & $2.664 \pm 2.107$ & $\mathbf{0.479 \pm 0.200}$ \\
 & AS & $0.467 \pm 0.090$ & $0.914 \pm 0.284$ & $0.956 \pm 0.126$ \\
 & SSS (raw) & $0.227 \pm 0.075$ & $0.924 \pm 0.313$ & $0.523 \pm 0.144$ \\
 & SSS (corrected) & $\underline{0.192 \pm 0.010}$ & $\underline{0.109 \pm 0.034}$ & $\underline{0.482 \pm 0.014}$ \\
 & SSS (buffer) & $0.194 \pm 0.010$ & $\mathbf{0.105 \pm 0.014}$ & $\mathbf{0.479 \pm 0.016}$ \\
\midrule
LJ-13 ($d=39$) & TA-BG & $1.21 \pm 0.02$ & $61.47 \pm 0.12$ & $4.16 \pm 0.01$ \\
 & PITA & $0.04 \pm 0.00$ & $2.26 \pm 0.21$ & $1.65 \pm 0.00$ \\
 & AS & $0.899 \pm 0.770$ & $\ast$ & $3.179 \pm 1.485$ \\
 & SSS (raw) & $0.041 \pm 0.032$ & $\ast$ & $\underline{1.499 \pm 0.043}$ \\
 & SSS (corrected) & $\underline{0.026 \pm 0.027}$ & $\underline{1.405 \pm 1.733}$ & $1.626 \pm 0.059$ \\
 & SSS (buffer) & $\mathbf{0.006 \pm 0.002}$ & $\mathbf{0.324 \pm 0.063}$ & $\mathbf{1.465 \pm 0.003}$ \\
\bottomrule\end{tabular}\end{lrbox}
\ifdim\wd\resultbox>\linewidth\resizebox{\linewidth}{!}{\usebox{\resultbox}}\else\usebox{\resultbox}\fi
\par\medskip
\caption{{Alanine Dipeptide at 300 K. Discrepancies and ranks follow Table~\ref{tab:reported}. Results for \PITA{} and \TABG{} are from \citet{akhoundsadegh2025progressive}.}}
\label{tab:ala}
\setlength{\tabcolsep}{5pt}
\begin{lrbox}{\resultbox}\begin{tabular}{@{}lrrr@{}}
\toprule
Method & Energy-$\mathcal W_1\downarrow$ & Energy-$\mathcal W_2\downarrow$ & $\mathbb{T}$-$\mathcal W_2\downarrow$ \\
\midrule
TA-BG & $83.457 \pm 0.070$ & $86.176 \pm 0.104$ & $0.979 \pm 0.012$ \\
PITA & $1.530 \pm 0.068$ & $1.615 \pm 0.053$ & $\underline{0.270 \pm 0.023}$ \\
AS & $\ast$ & $\ast$ & $1.490 \pm 0.003$ \\
SSS (raw) & $3.765 \pm 0.302$ & $3.858 \pm 0.305$ & $\mathbf{0.222 \pm 0.058}$ \\
SSS (corrected) & $\underline{1.230 \pm 1.422}$ & $\underline{1.432 \pm 1.525}$ & $0.757 \pm 0.745$ \\
SSS (buffer) & $\mathbf{0.212 \pm 0.190}$ & $\mathbf{0.257 \pm 0.181}$ & $0.399 \pm 0.207$ \\
\bottomrule\end{tabular}\end{lrbox}
\ifdim\wd\resultbox>\linewidth\resizebox{\linewidth}{!}{\usebox{\resultbox}}\else\usebox{\resultbox}\fi
\end{table}

{Table~\ref{tab:reported} compares the learned samplers on the Gaussian mixtures and particle systems. The three SSS rows use the same trained model for each target. Across the ten target--metric comparisons in this table, an SSS sample set attains the lowest displayed mean in nine cases, including one tie, and the second-lowest mean in the remaining case.}

\paragraph{Gaussian mixtures.}
{Corrected SSS samples have the lowest mean distance and energy errors on both mixtures. On GMM50, correction reduces Distance-$\mathcal W_2$ from $0.166$ to $0.048$ and Energy-$\mathcal W_2$ from $4.304$ to $0.321$.}

\paragraph{Particle systems.}
{On DW-4, \PITA{} has the lowest mean distance error. The SSS buffer has the lowest energy error and ties \PITA{} on geometric error at the displayed precision ($0.479$). Corrected SSS ranks second on all three DW-4 metrics. On LJ-13, the buffer has the lowest mean error on all three metrics. The raw SSS samples have a small geometric error, yet their energy error exceeds $10^4$.}

\paragraph{{Alanine Dipeptide.}}
{Table~\ref{tab:ala} compares the distribution of backbone angles and the distribution of energies. Raw SSS samples have the lowest mean angular Wasserstein distance, while the buffer ranks first on both energy metrics. Its Energy-$\mathcal W_2$ is $0.257$, followed by corrected SSS at $1.432$. Thus the best sample set depends on whether we measure backbone conformations or energies. See Appendix~\ref{app:costs} for the evaluation budgets.}

\paragraph{Choice of sample set.}
{A trained model can generate fresh samples whenever needed. The mixture results show how much a subsequent importance correction can improve those draws. The buffer offers a third option when a stored population suffices: on LJ-13 it has the lowest errors across all three metrics, although the number of distinct configurations is limited by its capacity. These differences motivate the next experiments, which examine how correction and local exploration shape the population used for training.}

\subsection{Ablation experiments}
\label{sec:ablation}
We now investigate different aspects of \method{} through a set of ablation studies. We run these experiments on DW-4 to examine the roles of importance-correction and MCMC rejuvenation. Specifically, we address two questions.

\textbf{Question 1 (Path-space weight resampling).} The diffusion can propose samples in different modes without assigning them the correct relative masses. Does importance resampling improve these masses, or are the retained MCMC updates sufficient?

\textbf{Question 2 (MCMC rejuvenation).} After resampling, MCMC explores nearby configurations and diversify duplicated particles. How does this exploration affect the training data and subsequent learning?

\begin{figure}[tbp]
\centering
\begin{subfigure}[t]{0.68\textwidth}
  \centering
  \includegraphics[width=\linewidth]{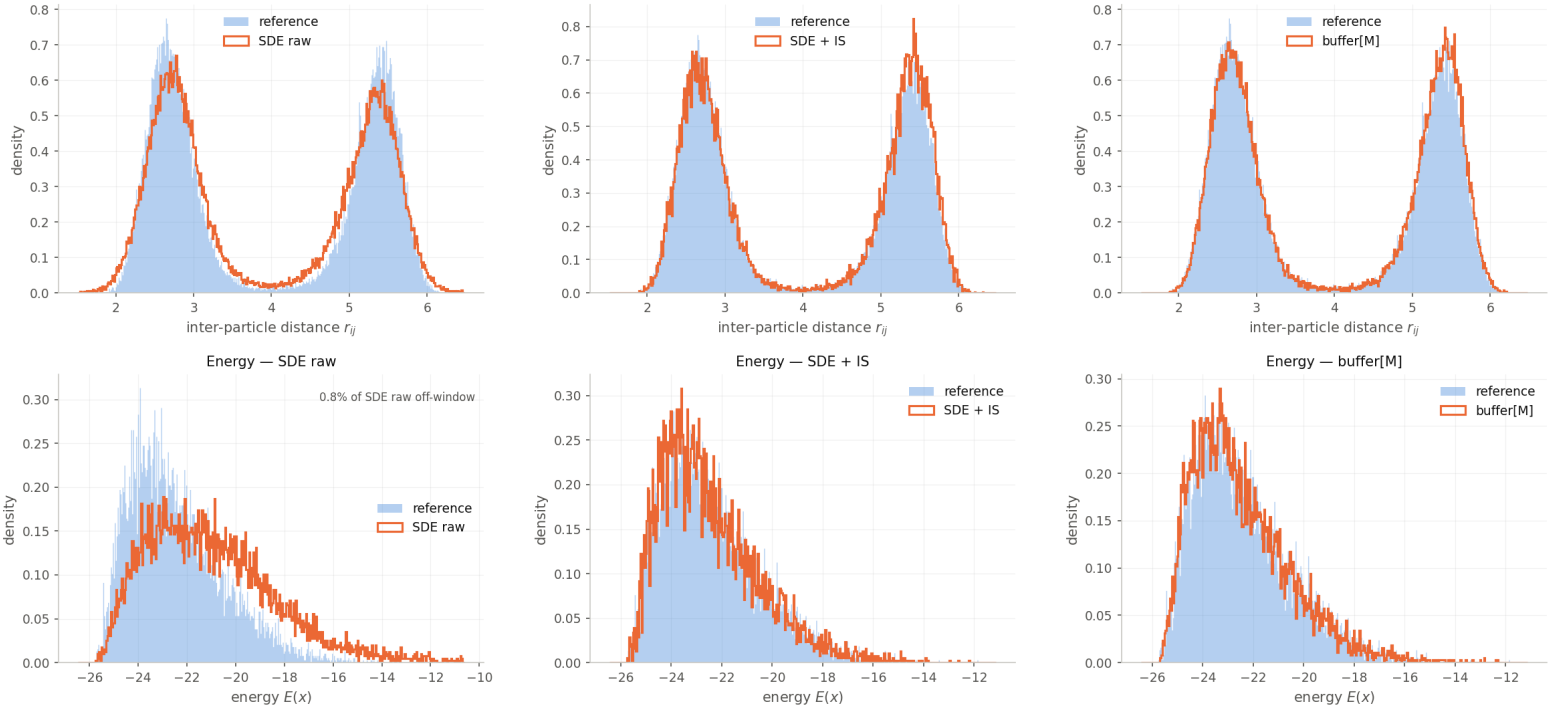}
  \caption{\method{} full method.}
  \label{fig:dw4-full}
\end{subfigure}

\vspace{0.5em}

\begin{subfigure}[t]{0.68\textwidth}
  \centering
  \includegraphics[width=\linewidth]{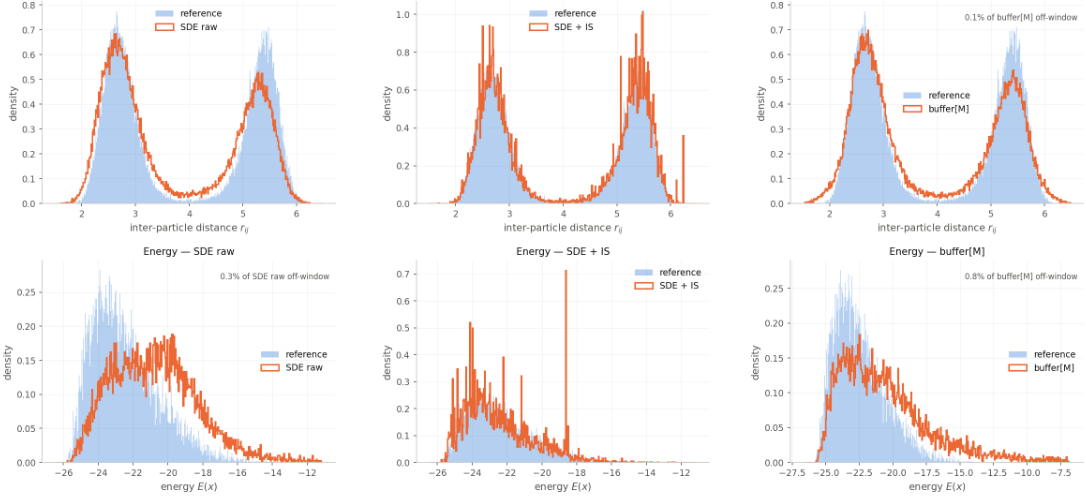}
  \caption{\method{} without path-space correction.}
  \label{fig:dw4-nocorrection}
\end{subfigure}

\caption{Ablation of the path-space correction on DW-4, final generated samples (orange) and ground truth (blue). Each panel shows the three sample sets of \method{} (from left to right): raw samples, importance-corrected samples, and the replay buffer. The top row shows the distribution of inter-particle distances $r_{ij}$ and the bottom row shows the energy distribution.}
\label{fig:ablation-correction}
\end{figure}

To answer Question 1, we compare \method{} with a variant that omits the path-space importance correction: it uses the same architecture, SMC initialization, HMC budget, temperature bridge, and training schedule. Only the resampling by the correction weights is removed when refreshing the buffers. Figure~\ref{fig:ablation-correction} compares the two runs on DW-4. Without path-space correction, both modes of the pairwise-distance distribution remain visible, but their relative masses are distorted. The same distortion appears in the final buffer, and the buffer also acquires a long tail of high-energy configurations. These batches are then reused for training. The normalized path-space ESS, computed from the weights in~\eqref{eq:path-weight} at $\beta=1$, {drops sharply when path-space correction is removed}. Thus, we conclude that the importance-correction, and not the MCMC updates, is what keeps the training data faithful to the relative masses of the target.

\begin{figure}[tbp]
\centering
\begin{subfigure}[t]{0.68\textwidth}
  \centering
  \includegraphics[width=\linewidth]{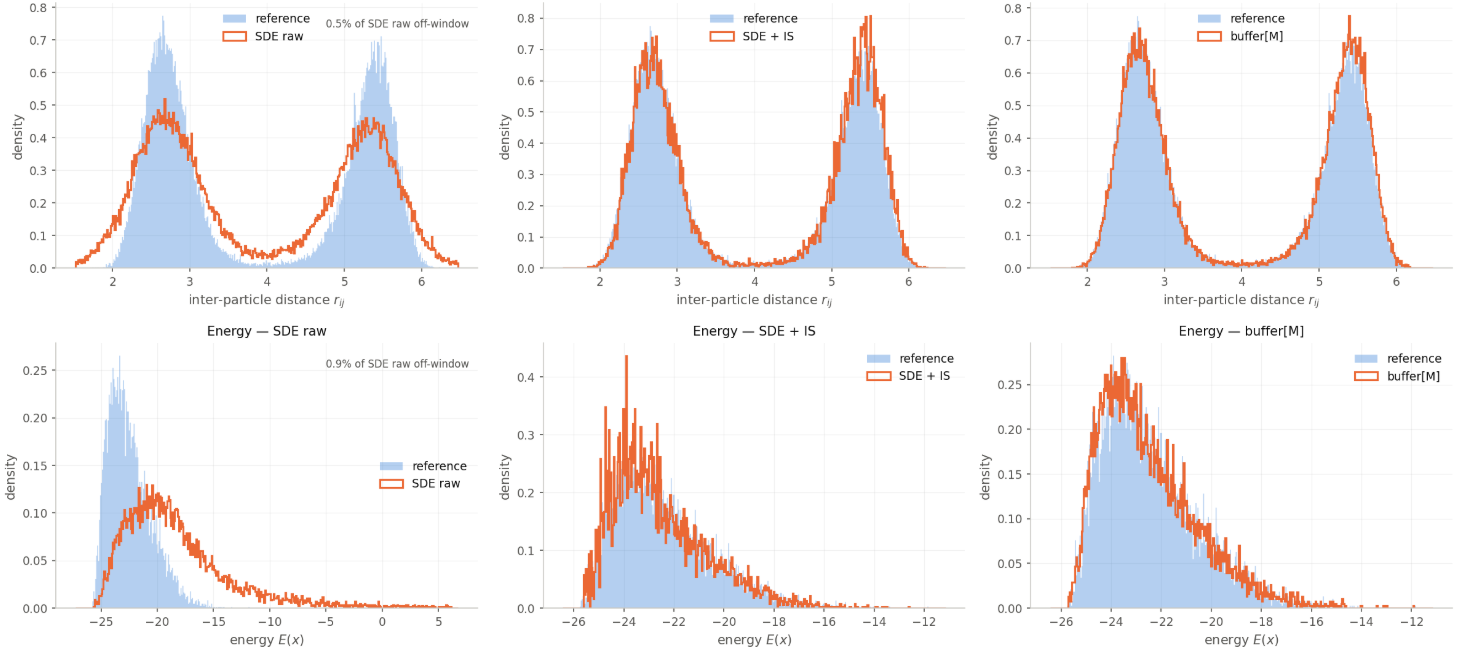}
  \caption{\method{} full method.}
  \label{fig:dw4-HMC-full}
\end{subfigure}

\vspace{0.5em}

\begin{subfigure}[t]{0.68\textwidth}
  \centering
  \includegraphics[width=\linewidth]{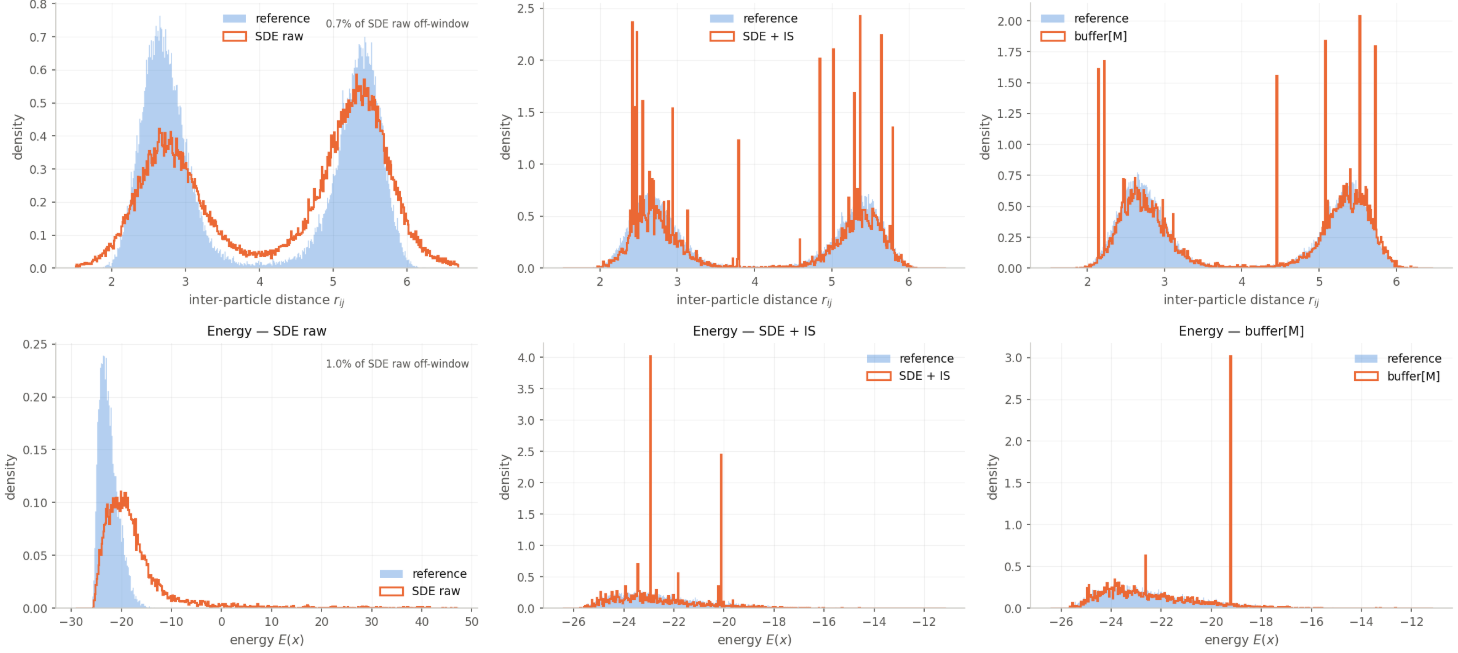}
  \caption{\method{} without HMC rejuvenation.}
  \label{fig:dw4-noHMC}
\end{subfigure}

\caption{Ablation of the HMC rejuvenation on DW-4, evaluated at $\beta=1$ midway through training (orange) and ground truth (blue). Each panel shows the three sample sets of \method{} (from left to right): raw samples, importance-corrected samples, and the replay buffer. The top row shows the distribution of inter-particle distances $r_{ij}$ and the bottom row the energy distribution.}
\label{fig:ablation-HMC}
\end{figure}

To answer Question~2, we compare SSS with a variant that omits the HMC steps. Both runs share the architecture, SMC initialization, path-space correction mechanism, temperature bridge, and training schedule. Only the rejuvenation step is removed when refreshing the buffers. Figure~\ref{fig:ablation-HMC} compares the two runs at the same intermediate training iteration. Without rejuvenation, the resampled batches consist of a few configurations repeated many times, which appear as sharp spikes in the distance and energy histograms. These spikes are inherited by the training buffer, so the model is trained on a small number of distinct configurations. Consequently, training is slower and less stable. Specifically, at the recorded iteration, the raw samples of the full method already reproduce both distance modes and a plausible energy distribution, whereas those of the variant remain far from the reference and carry a long tail of high-energy configurations. The normalized path-space ESS at $\beta = 1$ reflects the same gap, {dropping significantly} when rejuvenation is removed. These results indicate that MCMC exploration improves the diversity of the importance-resampled training populations and accelerates learning in this experiment.

{In all of our experiments, we keep the first positive-temperature buffer fixed, choose a particle count for SMC initialization, and reuse its HMC kernel during training. Appendix~\ref{app:additional} examines how sensitive the results are to each of these choices.}

\section{Related work}
\label{sec:related}

\paragraph{Learning proposals with Monte Carlo.}
Annealed Flow Transport and CRAFT combine
learned maps with sequential weighting and MCMC
\citep{arbel2021annealed,matthews2022continual}; FAB learns a reusable flow
from annealed-importance-sampling computations \citep{midgley2023flow}.
Reinforced SMC couples learned proposals and twist functions with SMC,
off-policy reinforcement learning, and replay
\citep{choi2025reinforced}. SCLD likewise combines controlled diffusions with
sequential resampling \citep{chen2025sequential}. Other methods also combine amortization and Monte Carlo correction
\citep{doucet2022score,vargas2024transport,albergo2025nets}. Our approach trains one sampler with temperature as an input, using supervised learning on repeatedly refreshed, corrected training data from all levels.

\paragraph{Learning across temperatures.}
PTSD trains diffusion samplers sequentially across temperatures using
parallel tempering \citep{swendsen1986replica,geyer1991markov,hukushima1996exchange,earl2005parallel} and MCMC refinement \citep{rissanen2025progressive}.
\TABG{} learns a flow through temperature reweighting
\citep{schopmans2025temperature}. \PITA{} combines progressive diffusion
training with inference-time annealing to produce lower-temperature data
\citep{akhoundsadegh2025progressive}. We share their use of easier targets
to train harder ones, an idea also exploited by temperature-conditioned generative models \citep{dibak2022temperature,wang2022data,invernizzi2022skipping}, but jointly train a shared model across all temperatures.

\paragraph{Score learning and weighted stochastic samplers.}
iDEM alternates generation and energy-based score learning
\citep{akhoundsadegh2024iterated}, while importance-weighted score matching
corrects the distribution used in a score-matching objective
\citep{wang2025importance}. {Adjoint Sampling derives an endpoint-gradient regression from a stochastic control problem \citep{havens2025adjoint}. Its training distribution changes with the current control; our regression approximates a prescribed tempered distribution at each level.} Our relative-score target uses known Gaussian
identities \citep{efron2011tweedie,vincent2011connection,phillips2024particle,debortoli2024target}. Other weighting construction in the literature include Boltzmann generators
combining learned equilibrium proposals with reweighting
\citep{noe2019boltzmann,klein2023equivariant,tan2025scalable}, and stochastic normalizing flows computing trajectory
weights without marginalizing stochastic transitions
\citep{wu2020stochastic}, building on nonequilibrium path-space reweighting \citep{jarzynski1997nonequilibrium,crooks1999entropy,delmoral2006smc,nilmeier2011nonequilibrium}. %

\section{{Discussion and Limitation}}

\paragraph{{Contributions.}}
{
Our main contribution is a corrected self-training scheme
that jointly improves a shared proposal and its training
populations across a temperature ladder. The individual
ingredients are established, but their combination differs
from existing approaches. Unlike progressive training
methods such as \PITA{}, which generate training data as
temperatures are visited, our scheme repeatedly refreshes
corrected populations at multiple temperatures while
training one model jointly across the ladder. It also
differs from flow-based bootstrapping and learned transport:
the model is fitted to weighted populations approximating
the tempered targets, rather than learning transport maps
between them. At each refresh, proposals are corrected
and rejuvenated at one temperature, then reweighted to
update the buffer at the next. This allows information
collected at one temperature to improve proposals at
others, with no need to differentiate through sampling
or correction.

Our second contribution is Sequential Score Sampler
(\method{}), a diffusion implementation of this idea.
A single network learns relative scores across temperatures
by regression on the corrected populations. The
Gaussian transitions provide computable path-space
importance weights, allowing proposals to be corrected
without evaluating their endpoint density or learning
an auxiliary density model. The same training and correction
can be used with different proposal families and
fitting objectives.

All in all, this leads to a workflow that is simple to explain and to implement and that has demonstrably competitive performance. 
}

\paragraph{{Performance.}}
{
Our experiments evaluate both the benefit of corrected self-training and the performance of the resulting sampler. On GMM50, repeated buffer refreshing reduces the mean distance discrepancy of raw samples from 0.398 with fixed SMC training data to 0.166. Among the compared learned samplers, corrected SSS samples have the lowest mean distance and energy discrepancies on both Gaussian mixtures. On DW-4, the buffer has the lowest mean energy error and ties \PITA{} on geometric error at the displayed precision. On LJ-13, the buffer leads on all three metrics. On Alanine Dipeptide, raw samples give the lowest mean backbone-angle discrepancy, while the buffer has the lowest mean Energy-$\mathcal W_1$ and
Energy-$\mathcal W_2$. At a matched energy budget, long-run
SMC is competitive on DW-4 and slightly more accurate on
LJ-13. This highlights a trade-off: continued SMC can produce
more accurate populations at a fixed budget, whereas training
also produces a reusable proposal for generating fresh samples.
}

\paragraph{{Limitations.}}
{
The method depends on adequate coverage of the target
distributions by the training buffers. If $\Data_k$ misses
a mode, resampling cannot recover it, and short MCMC runs
may fail to discover it. The next buffer can then inherit
the gap, which repeated training may reinforce.
Even when all modes are represented, concentrated importance
weights can lead to low effective sample size and many
duplicates after resampling. Local MCMC moves can diversify
these copies but may not restore the relative masses of
distant modes. Self-normalized importance estimates also
have finite-sample bias. Finally, initialization and repeated
buffer refreshes incur computational costs, so the value
of amortization depends on subsequent reuse of the learned
sampler. Appendix~\ref{app:engineering} discusses the
implications for initialization and rejuvenation.
}

\label{sec:main-end}

\section*{Reproducibility statement}
The method is specified by Algorithms~\ref{alg:train}-\ref{alg:sample} and Equations~\eqref{eq:loss}-\eqref{eq:path-weight}, with algorithmic details in Appendix~\ref{app:implementation} and a justification of the score formula in Appendix~\ref{app:theory}. Appendix~\ref{app:experimental} gives the target definitions, network architectures, evaluation metrics, numerical settings, and the description on re-implementation of the baselines. {Table~\ref{tab:time} reports timings on a single NVIDIA RTX 4500 Ada Generation GPU. The LJ-13 and Alanine Dipeptide results for \PITA{} and \TABG{} come from \citet{akhoundsadegh2025progressive}.} Code to reproduce our method is available at \url{https://github.com/VietNH-researching/progressive_is_tempering.git}.
\setlength{\bibsep}{3pt plus 1pt minus 1pt}
\bibliographystyle{plainnat}
\bibliography{ref_cleaned}

\newpage
\appendix
\numberwithin{equation}{section}
\part*{Appendix}
\addcontentsline{toc}{part}{Appendix}

\etocsettocstyle{\section*{Contents}}{}
\localtableofcontents

\section{Algorithmic details}
\label{app:implementation}

\subsection{Adaptive SMC initialization}
\label{app:smc}
Starting from $N_{\mathrm{SMC}}$ independent draws from $\rho$ with uniform weights, adaptive SMC chooses each next temperature so that the incremental importance ESS equals a prescribed fraction $\eta$ of the population size. Specifically, at step $k$, reweighting the equally weighted population $\{x_i^{(k)}\}$ to a candidate $\beta>\beta_k$ gives the incremental log weights $-(\beta-\beta_k)E(x_i^{(k)})$ and the normalized ESS
\begin{equation}
\mathrm{ESS}_k(\beta)=\frac{\big(\sum_i w_i(\beta)\big)^2}{N_{\mathrm{SMC}}\sum_i w_i(\beta)^2},\qquad w_i(\beta)=\exp\{-(\beta-\beta_k)E(x_i^{(k)})\}.
\label{eq:local-ess}
\end{equation}
If $\mathrm{ESS}_k(1)\ge\eta$, we set $\beta_{k+1}=1$ and the sequence ends. Otherwise, $\beta_{k+1}$ is the root of $\mathrm{ESS}_k(\beta)=\eta$ on $[\beta_k,1]$. Since $\mathrm{ESS}_k$ is non-increasing in $\beta$, this root can be found by bisection. We take the left end of the final bisection bracket, where $\mathrm{ESS}_k\ge\eta$, so each step is slightly conservative. The number of temperatures $K$ is therefore determined by $\eta$, $N_{\mathrm{SMC}}$ and the target, not set in advance.

The population is then resampled by systematic resampling, applied after a uniformly random permutation of the particles, and moved by MCMC targeting the newly selected distribution $\pi_{\beta_{k+1}}$. This is the standard adaptive resample-move construction \citep{delmoral2006smc,jasra2011adaptive,gilks2001following}.

We use preconditioned one-step HMC with full momentum refresh as our MCMC kernel. Let $\widehat\Sigma_k$ be the symmetrized empirical covariance of the resampled particles at $\beta_k$ and $L_k$ its Cholesky factor. If $\widehat\Sigma_k$ is not positive definite, we use $L_k=\operatorname{diag}(\widehat\Sigma_k)^{1/2}$ instead. Given a step size $\epsilon_k$, one move from $x$ is
\begin{equation}
\begin{alignedat}{2}
  &v \sim \mathcal N(0,I_d),
  &\qquad &x_{1/2} = x + \frac{\epsilon_k}{2}L_k v, \\
  &v' = v + \epsilon_k L_k^\top \nabla\log\pi_{\beta_k}(x_{1/2}),
  &\qquad &x' = x_{1/2} + \frac{\epsilon_k}{2}L_k v'.
\end{alignedat}
\label{eq:hmc}
\end{equation}
where $\nabla\log\pi_{\beta_k}=\nabla\log\rho-\beta_k\nabla E$. The move is accepted with probability
\begin{equation}
\alpha(x,x')=\min\Big\{1,\exp\Big(\log\pi_{\beta_k}(x')-\tfrac12\|v'\|^2-\log\pi_{\beta_k}(x)+\tfrac12\|v\|^2\Big)\Big\}.
\label{eq:hmc-accept}
\end{equation}
For fixed $L_k$, this is ordinary HMC after the linear change of coordinates $x\mapsto L_k^{-1}x$; the Metropolis acceptance rule preserves $\pi_{\beta_k}$. The step size $\epsilon_k$ is chosen by bisection so that the mean acceptance probability of one trial move over the population lies in $[0.40,0.65]$. Each particle then takes $n_{\mathrm{MCMC}}$ moves (specified in Appendix~\ref{app:hyperparameters}). The adapted parameters $(\epsilon_k,L_k)$ are then frozen and reused for rejuvenation at $\beta_k$ during self-training. Each subsequent kernel therefore leaves its target exactly invariant, although invariance alone does not guarantee rapid inter-basin mixing.

The populations obtained at $\beta_1,\dots,\beta_K$ approximate the corresponding targets and initialize $\Data_1,\dots,\Data_K$ with equal weights; $\Data_0$ is filled with independent draws from $\rho$. For DW-4 and LJ-13, particles are represented in ambient coordinates and the retained populations are centered before entering the replay buffers. Appendix~\ref{sec:com} explains the centered target for each system and the corresponding sampling convention.

\subsection{Replay buffers and per-batch weight normalization}
\label{app:replay}

Each buffer has capacity $C=N_{\mathrm{SMC}}$. $\Data_0$ contains
independent draws from $\rho$; $\Data_1, \ldots, \Data_K$ are initialized with the retained SMC populations. During training, $\Data_0$ and $\Data_1$ remain
fixed, other buffers are updated every iteration. Specifically, at each active temperature $\beta_k$, $k = 1, \ldots, K-1$, we generate $N$ new proposal samples per
refresh, correct them with importance resampling, rejuvenate with HMC, and insert the resulting weighted samples into $\Data_{k+1}$. Thus, only
$\Data_2,\ldots,\Data_K$ have their initial samples removed through first-in, first-out (FIFO) replacement. They are fully replaced after $\lceil C/N\rceil$ refreshes, subject to the convention for removing partially retained batches.

After resampling according to the proposal importance weights and applying MCMC at $\beta_k$, let
$\ell_i=-(\beta_{k+1}-\beta_k)E(x_i)$ be the bridge log weights. Store
\begin{equation}
  \widetilde\ell_i
  =\ell_i-\lse(\ell_1,\ldots,\ell_N)+\log N.
  \label{eq:block-normalization}
\end{equation}
Here $\lse(\ell_1,\ldots,\ell_N)=\log\sum_i e^{\ell_i}$. Each batch has total weight $N$, matching an equally sized batch of initial samples with unit weights, while its relative weights are unchanged. Thus each refresh contributes a self-normalized approximation of the target. Without this normalization, fluctuations in the batch estimate of $Z_{\beta_{k+1}}/Z_{\beta_k}$ would also determine its influence relative to other batches.

We use systematic resampling both to correct proposals and to draw training minibatches. The resulting samples can be dependent.
Per-batch normalization and additional optimization do not make the finite adaptive buffer an exact sample from its target, and we do not claim convergence of the self-training loop at fixed buffer size.

At each optimization step, we draw a minibatch of $b$ samples from each buffer $\mathcal{D}_k$ by systematic resampling according to the stored weights, sample fresh diffusion times {$t$ from the schedule for the target} and perturbations $\epsilon\sim\mathcal{N}(0,I_d)$, and take one Adam step on~\eqref{eq:loss}. We perform $J$ such steps between consecutive buffer refreshes. Energy gradients are computed once per stored sample and cached; no gradients pass through the procedure that generates or selects the samples. The value of $J$, the number of outer iterations and the optimizer settings are listed in Appendix~\ref{app:hyperparameters}.

\subsection{Numerical generation and stable log weights}
\label{app:numerics}

The path weight can be accumulated during sampling without storing the trajectory. Write $h=T/L$, $a=e^{-h}$, $\sigma=\sqrt{1-a^2}$, and
$c_n=2(1-a)r_\theta(y_n,T-nh,\beta)$. One step is
\[
  y_{n+1}=ay_n+c_n+\sigma\epsilon_n,
  \qquad \epsilon_n\sim\Normal(0,I_d).
\]
The proposal and reference transitions have the same covariance, so their Gaussian normalizers cancel. The log-weight increment is simply
\begin{equation*}
\begin{aligned}
  \Delta_n
  &=\log\frac{g_h(y_{n+1}\mid y_n)}
                 {q_{\theta,n}(y_{n+1}\mid y_n,\beta)}
  =\frac{\norm{y_{n+1}-ay_n-c_n}^2-
           \norm{y_{n+1}-ay_n}^2}{2\sigma^2}
  =-\frac{c_n^\top\epsilon_n}{\sigma}
    -\frac{\norm{c_n}^2}{2\sigma^2}.
\end{aligned}
\end{equation*}
The last expression uses the noise already drawn for the step and avoids subtracting two squared norms. Starting from zero, accumulate these increments and add the terminal energy contribution:
\begin{equation}
  \log w=\sum_{n=0}^{L-1}\Delta_n-\beta E(y_L).
  \label{eq:path-weight-impl}
\end{equation}
This is algebraically the same weight as~\eqref{eq:path-weight}. When $r_\theta\equiv0$, every increment is zero; at $\beta=0$, the full log weight is zero for any step size.

\begin{algorithm}[h]
\caption{Gaussian-reference sampler with online log weights}
\label{alg:sample}
\begin{enumerate}[label=\arabic*:,nosep,leftmargin=2em]
\item[] \textbf{Input:} $r_\theta$, energy $E$, temperature $\beta$, horizon $T$, steps $L$
\item $h\gets T/L$, $a\gets e^{-h}$, $\sigma\gets\sqrt{1-e^{-2h}}$
\item $y\sim\Normal(0,I_d)$, $\ell\gets0$
\item \textbf{for} $n=0,\ldots,L-1$ \textbf{do}
\item \hspace*{1em}$c\gets2(1-a)\,r_\theta(y,T-nh,\beta)$; $\epsilon\sim\Normal(0,I_d)$
\item \hspace*{1em}$y'\gets ay+c+\sigma\epsilon$
\item \hspace*{1em}$\ell\gets\ell-c^\top\epsilon/\sigma-\norm{c}^2/(2\sigma^2)$
\item \hspace*{1em}$y\gets y'$
\item $\ell\gets\ell-\beta E(y)$
\item \textbf{return} $(y,\ell)$
\end{enumerate}
\end{algorithm}

Weights are normalized in log space. At positive $\beta$, the terminal energy term downweights high-energy endpoints; an energy evaluated as $+\infty$ gives zero weight when the transition terms are finite. Resampling requires at least one finite log weight, and a batch with none must be treated as a failed refresh rather than normalized. This numerical issue is separate from the validity of the exact weight formula.

\section{Justification of the score formula}
\label{app:theory}

\paragraph{The regression target.}
Fix $\beta$ and let
$k_t(z\mid x)=\Normal(z;\alpha_t x,(1-\alpha_t^2)I_d)$, with $\alpha_t=e^{-t}$.
Assume that $E$ is differentiable and that differentiation under the integrals and the integration by parts below are valid. The Gaussian kernel satisfies
$\rho(x)k_t(z\mid x)=\rho(z)k_t(x\mid z)$.
Consequently, for $X\sim\pi_\beta$ and $X_t\mid X\sim k_t(\cdot\mid X)$,
\begin{align*}
  \frac{\pi_{\beta,t}(z)}{\rho(z)}
  =\frac{1}{Z_\beta}\int e^{-\beta E(x)}k_t(x\mid z)\,\dd x.
\end{align*}
Since $\nabla_z k_t(x\mid z)=-\alpha_t\nabla_x k_t(x\mid z)$, integration by parts gives
\begin{align*}
  \nabla_z\log\frac{\pi_{\beta,t}(z)}{\rho(z)}
  &=\alpha_t\,
    \frac{\int \nabla_x e^{-\beta E(x)}k_t(x\mid z)\,\dd x}
         {\int e^{-\beta E(x)}k_t(x\mid z)\,\dd x}
  =-\alpha_t\beta\,\E[\nabla E(X)\mid X_t=z].
\end{align*}
The final equality follows because the conditional density of $X$ given $X_t=z$ is proportional to $e^{-\beta E(x)}k_t(x\mid z)$. This is the relative-score identity used in~\eqref{eq:relative-score} \citep{debortoli2024target,phillips2024particle}.
Least-squares regression learns the conditional mean of its target. Thus, when $\E_{\pi_\beta}\norm{\nabla E(X)}^2<\infty$ for $\beta>0$, the population minimizer with regression target $-\alpha_t\beta\nabla E(X)$ is $r^*(X_t,t,\beta)$. At $\beta=0$, both are zero. Finite-buffer sampling and the target clipping used in the particle experiments are approximations to this population objective, not part of the identity.

\paragraph{The reverse sampler.}
The forward Ornstein--Uhlenbeck process has drift $-x$ and noise coefficient $\sqrt{2}$. Its standard reverse-time formula \citep{song2021score}, written with sampling time $s$ increasing from $0$ to $T$, has drift
\begin{align*}
  x+2\nabla_x\log\pi_{\beta,T-s}(x)
  =-x+2r^*(x,T-s,\beta).
\end{align*}
With the exact score and initial distribution $\pi_{\beta,T}$, this process ends at $\pi_\beta$. The explicit perturbation
$X_T=\alpha_T X+\sqrt{1-\alpha_T^2}\,\epsilon$
approaches the Gaussian reference as $T$ increases, motivating initialization from $\rho$. At finite $T$, this replacement is an approximation. The discrete importance weights in~\eqref{eq:path-weight} remain valid for the numerical sampler regardless of this initial mismatch or score error.

\section{Experimental details}
\label{app:experimental}
\subsection{Target definitions and reference distributions}
\label{app:target-definitions}

\paragraph{Gaussian mixture model (GMM).}
We consider a synthetic target on $\mathbb{R}^{d}$ with independent
coordinates, each following a two-component Gaussian mixture:
\begin{align*}
    \pi(x)
    =
    \prod_{j=1}^{d}
    \left[
        w_1 \mathcal{N}(x_j;\mu_1,\sigma^2)
        +
        w_2 \mathcal{N}(x_j;\mu_2,\sigma^2)
    \right],
\end{align*}
where $(\mu_1,\mu_2)=(-1.5,2.5)$, $(w_1,w_2)=(0.2,0.8)$,
and $\sigma=0.5$. We use $d=16$ and $d=50$, keeping the per-coordinate
parameters fixed so that the two settings differ only in dimension.
This construction yields $2^{d}$ well-separated modes with unequal
probability masses. The main challenge is to explore these modes and
recover their relative masses despite the low-density regions separating
them. Exact independent samples can be obtained by sampling the mixture
component and Gaussian noise independently for each coordinate.

\paragraph{4-particle double-well (DW-4).}
DW-4 is an interacting-particle benchmark for sampling multimodal
Boltzmann distributions \citep{kohler2020equivariant}.
A configuration consists of four particles
$x^1,\ldots,x^4\in\mathbb{R}^2$, represented by $x\in\mathbb{R}^8$.
The target density is
\begin{align*}
    \pi(x) \propto \exp\!\left[-\tau U_{\mathrm{DW}}(x)\right], \quad 
    U_{\mathrm{DW}}(x)
    =
    \sum_{1\leq i<j\leq 4}
    \left[
        a(r_{ij}-r_0)^4
        +
        b(r_{ij}-r_0)^2
        +
        c
    \right],
\end{align*}
where $r_{ij}=\|x^i-x^j\|_2$ and $\tau$ is the physical inverse
temperature. We set $a=0.9$, $b=-4$, $c=0$, $r_0=4$, and $\tau=1$.
Since the energy is translation-invariant, we impose
$\sum_{i=1}^{4}x^i=0$ and define the density with respect to Lebesgue
measure on this six-dimensional subspace.
The two preferred pairwise distances give rise to competing particle
arrangements and multiple energy minima. The main sampling challenge
is to cross the energy barriers separating these arrangements. As reference samples, we use the MCMC samples of \citet{klein2023equivariant}, obtained from the DEM repository of \citet{akhoundsadegh2024iterated}.

\paragraph{13-particle Lennard--Jones (LJ-13).}
LJ-13 models a cluster of thirteen particles
$x^1,\ldots,x^{13}\in\mathbb{R}^3$,
with configurations represented by $x\in\mathbb{R}^{39}$.
The Lennard--Jones potential models short-range repulsion and
longer-range attraction between particles and provides a standard
benchmark for sampling atomic cluster configurations
\citep{kohler2020equivariant}.
We use the confined, regularized target
\begin{align*}
    \pi(x) \propto \exp\!\left[-\tau U_{\mathrm{LJ}}(x)\right], \quad
    U_{\mathrm{LJ}}(x)
    =
    \epsilon
    \sum_{1\leq i<j\leq 13}
    \left[
        \left(\frac{r_m}{\widetilde r_{ij}}\right)^{12}
        -
        2\left(\frac{r_m}{\widetilde r_{ij}}\right)^6
    \right]
    +
    \frac{\kappa}{2}\sum_{i=1}^{13}\|x^i\|_2^2,
\end{align*}
where $\widetilde r_{ij} = \sqrt{\|x^i-x^j\|_2^2+\delta}$. We set $\epsilon=2$, $r_m=1$, $\kappa=1$, $\delta=10^{-5}$,
and the physical inverse temperature $\tau=1$.
The harmonic term confines the cluster, while $\delta$ regularizes
the repulsive interaction near particle collisions.
We impose $\sum_{i=1}^{13}x^i=0$ and define the density with respect
to Lebesgue measure on the resulting 36-dimensional subspace.
The main sampling difficulties are transitions between metastable
cluster configurations and the steep energy gradients associated
with short-range repulsion. As for DW-4, the reference samples are the MCMC samples of \citet{klein2023equivariant}, obtained from the DEM repository of \citet{akhoundsadegh2024iterated}.

For the particle targets and the Gaussian reference, the relative energy used by the sampler is $E(x)=\tau U(x)-\norm{x}^2/2$, up to an additive constant, with $\tau=1$ in these experiments.

\paragraph{{Alanine Dipeptide.}}
{We sample the 22-atom ACE--ALA--NME molecule at $300$ K in implicit solvent, using the AMBER ff14SB force field and OBC1 solvent model of \citet{akhoundsadegh2025progressive}. The target density is proportional to $\exp[-U(x)/(k_BT_{\rm phys})]$ in the L-alanine sector. Its backbone angles $(\phi,\psi)$ distinguish conformations with different equilibrium populations. We compare with the $10{,}000$ frames in the 300 K molecular-dynamics test split used by \citet{akhoundsadegh2025progressive}.

SMC and HMC operate in 60 internal coordinates: 21 bonds, 20 angles and 19 torsions. The reference has 42 standard Gaussian coordinates for transformed bonds, angles and the chirality-defining improper torsion, and 18 independent uniform circular coordinates for the remaining torsions. The target density in these coordinates includes the coordinate Jacobian and the conformational volume factor. We wrap circular coordinates after each HMC position update. For diffusion training, we reconstruct Cartesian positions, subtract their center of mass, apply a random rotation and divide by the coordinate standard deviation of the final SMC population. The tempering reference in Cartesian space is the distribution induced by this map. The forward diffusion still uses Gaussian noise. Writing $\gamma$ for the standard Gaussian density on the centered Cartesian subspace, the path weight in~\eqref{eq:path-weight} uses the endpoint factor $\bar\pi_\beta(y_L)/\gamma(y_L)$ in scaled Cartesian coordinates. This reduces to $e^{-\beta E(y_L)}$ when $\rho=\gamma$.}

\subsection{Neural network architecture}
\label{app:architecture}
All experiments use a single network $r_\theta(x_t,t,\beta)$ shared across the whole temperature ladder, conditioned on both the diffusion time $t$ and the temperature parameter $\beta$. The architecture is chosen to match the structure of each target. We use a residual MLP for the Gaussian mixtures and an equivariant graph network {for DW-4 and LJ-13, and a diffusion transformer for Alanine Dipeptide.} On centered {DW-4 and LJ-13} configurations, the target is unchanged by permutations of identical particles, rotations, and reflections.
\paragraph{Gaussian mixtures.}
The relative score network $r_\theta(x_t,t,\beta)$ is a residual MLP with input $[x_t,\allowbreak\ \mathrm{emb}_{16}(t),\allowbreak\ \mathrm{emb}_{8}(\beta)]$, where $\mathrm{emb}_m$ denotes an $m$-dimensional sinusoidal embedding with log-spaced frequencies in $[1,10]$ for $t$ and $[0.5,5]$ for $\beta$. A linear layer maps the input to width $256$, followed by three pre-norm residual blocks (RMSNorm, Dense, ELU, Dense) and a linear output layer. The output weights are initialized from $\mathcal{N}(0,\sigma^2)$ with $\sigma^2 = 10^{-8}$, and the output bias is set to zero, so that $r_\theta\approx 0$ at initialization, consistent with the zero relative score at $\beta=0$ (Section~\ref{sec:regression}). The network has $4.1\times10^{5}$ parameters for $d=16$ and $4.3\times10^{5}$ for $d=50$.
\paragraph{Particle systems.}
For DW-4 and LJ-13, $r_\theta$ is an E($n$)-equivariant graph neural network (EGNN) \citep{satorras2021en} on the fully connected particle graph, with $5$ layers of width $128$, SiLU activations, attention-gated messages, and residual node updates. Each node receives $t$ as a scalar feature together with an $8$-dimensional sinusoidal embedding of $\beta$ (frequencies in $[0.5,5]$), and each edge carries the squared inter-particle distance of the input configuration. Coordinate updates are bounded by $\tanh$, with a per-layer range of $3$. The output is the total coordinate displacement with its mean removed. It is therefore equivariant to permutations, rotations, and reflections, invariant to translations, and lies in the zero center-of-mass subspace (Appendix~\ref{sec:com}). The network has $5.8\times10^{5}$ parameters.

\paragraph{{Alanine Dipeptide.}}
{We use the Cartesian diffusion transformer of \citet{akhoundsadegh2025progressive}. A linear layer embeds each atom's three coordinates into a token of width 192. Six transformer blocks with six attention heads are conditioned on the noise level $\tfrac18\log\sigma_t^2$ and inverse temperature $\beta$ through a 64-dimensional embedding. The output layer is initialized to zero. We scale its prediction by $\alpha_t/\sigma_t$, remove the center of mass and clip each component to $[-10^4,10^4]$. The network has $3{,}179{,}267$ parameters. Training uses dropout $0.1$ and fresh random rotations applied jointly to each configuration and its stored gradient. For the TSM term, we clip the clean relative score per atom at the 99.9th percentile of its norms in the initial buffer at that temperature.}

\subsection{Projection onto the zero center-of-mass subspace}
\label{sec:com}

For DW-4 and LJ-13, we represent a configuration of $k$ particles in
$\mathbb{R}^s$ as $y=(y^1,\ldots,y^k)\in\mathbb{R}^{ks}$.
We remove its center of mass by subtracting the mean particle position:
\begin{equation}
  x^i = y^i-\frac{1}{k}\sum_{j=1}^k y^j,
  \qquad i=1,\ldots,k.
\end{equation}
Equivalently, $x=\mathcal{A}y$, where
\begin{equation}
  \mathcal{A}
  = \left(I_k-\frac{1}{k}\mathbf{1}_k\mathbf{1}_k^\top\right)
    \otimes I_s
  \label{eq:com-projector}
\end{equation}
is the orthogonal projection onto
$\mathcal{X}^{\mathrm{CoM}}
=\{x\in\mathbb{R}^{ks}:\sum_i x^i=0\}$.
This subspace has dimension $(k-1)s$. All diffusion sampling and density calculations below are understood on this space. The Gaussian reference is standard in an orthonormal coordinate system on the subspace; in ambient coordinates its covariance is $\mathcal A$.

For DW-4, the physical energy depends only on pairwise distances. Fixing the center of mass removes the unconstrained translation and makes the target a distribution over internal configurations.
For LJ-13, the pair interaction is also translation-invariant, but the full potential is not, because it includes harmonic confinement. The confinement separates as
\[
  \sum_{i=1}^k \|y^i\|^2
  =
  \sum_{i=1}^k \|y^i-\bar y\|^2+k\|\bar y\|^2,
  \qquad
  \bar y=\frac{1}{k}\sum_{i=1}^k y^i.
\]
Thus the LJ-13 density factors into a center-of-mass term and an internal-coordinate term. Centering a draw preserves its internal-coordinate distribution; it does not assert translation invariance of the full confined target.

\paragraph{HMC steps.}
We use preconditioned HMC during initialization and training. Retained SMC particles are centered before entering the replay buffers; during training, the same centering is applied after each HMC move. The target-invariance claim requires the proposal and acceptance calculation to be consistent with the centered configuration space; projection alone does not establish it.

\paragraph{Diffusion steps.}
We center the Gaussian noise used in both training and sampling. For a centered buffer sample $x_0$, the perturbation is
\[
  x_t=\alpha_t x_0+\sqrt{1-\alpha_t^2}\,\mathcal A\epsilon,
  \qquad \epsilon\sim\Normal(0,I_{ks}).
\]
The score network also returns a centered vector, so sampling takes the form
\[
  y_{n+1}=ay_n+2(1-a)r_\theta(y_n,T-nh,\beta)
              +\sqrt{1-a^2}\,\mathcal A\epsilon_n.
\]
Starting from a centered reference draw, all states remain centered. The learned and reference transitions have the same covariance $(1-a^2)\mathcal A$ in ambient coordinates, or $(1-a^2)I$ in coordinates on the subspace. Their Gaussian normalizers therefore cancel, and the log-weight formula in Appendix~\ref{app:numerics} applies with the projected noise.

\subsection{Reported metrics}
\label{sec:reported-metrics}

\subsubsection{{Gaussian mixture models and Particle systems}}
{For the Gaussian mixtures, DW-4, and LJ-13, we use the following discrepancies:}
Distance-$\mathcal{W}_2$, Energy-$\mathcal{W}_2$, and
Geometric-$\mathcal{W}_2$, as considered in the evaluation of
PITA~\citep{akhoundsadegh2025progressive}. Lower values indicate better agreement with the reference samples.

\paragraph{Distance-$\mathcal{W}_2$.}
Let $h(x)\in\mathbb{R}^{f}$ denote a vector of summary statistics.
For the Gaussian mixture targets, $h(x)$ consists of the coordinates
of $x$; for DW-4 and LJ-13, it consists of all pairwise interparticle
distances. Given generated samples $\{x_i\}_{i=1}^{n}$ and reference
samples $\{y_j\}_{j=1}^{m}$, we pool these statistics into the
one-dimensional empirical distributions
\[
\widehat{\mu}_h
=
\frac{1}{nf}\sum_{i=1}^{n}\sum_{\ell=1}^{f}\delta_{h_\ell(x_i)},
\qquad
\widehat{\nu}_h
=
\frac{1}{mf}\sum_{j=1}^{m}\sum_{\ell=1}^{f}\delta_{h_\ell(y_j)},
\]
and report
\[
\text{Distance-}\mathcal{W}_2
=
\mathcal{W}_2(\widehat{\mu}_h,\widehat{\nu}_h).
\]
The one-dimensional $2$-Wasserstein distance admits the quantile
representation~\citep{peyre2019computational}
\[
\mathcal{W}_2(\mu,\nu)
=
\left(
\int_0^1
\left|F_\mu^{-1}(u)-F_\nu^{-1}(u)\right|^2\,\mathrm{d}u
\right)^{1/2},
\]
where $F_\mu^{-1}$ and $F_\nu^{-1}$ are the corresponding quantile
functions. Distance-$\mathcal{W}_2$ measures agreement in the pooled
coordinate or distance distribution, but does not capture dependencies
among statistics within each configuration.

\paragraph{Energy-$\mathcal{W}_2$.}
Following the energy-distribution evaluation used
by~\citet{akhoundsadegh2025progressive}, we compare the empirical distributions
of the energy $E(x)$:
\[
\text{Energy-}\mathcal{W}_2
=
\mathcal{W}_2\left(
\frac{1}{n}\sum_{i=1}^{n}\delta_{E(x_i)},
\frac{1}{m}\sum_{j=1}^{m}\delta_{E(y_j)}
\right).
\]
This metric measures whether generated samples reproduce the reference
energy distribution, complementing Distance-$\mathcal{W}_2$.
For both one-dimensional metrics, equal-sized scalar sample sets are
compared exactly by sorting and computing the root mean squared
difference. For unequal sizes, we approximate the quantile integral
using $1000$ uniformly spaced midpoint quantiles. The current evaluation excludes non-finite scalar values before either calculation. These metrics therefore do not penalize numerical failures: when exclusions occur, they describe only the retained finite subset and can understate the error of the full sample set.

\paragraph{Geometric-$\mathcal{W}_2$.}
Distance-$\mathcal{W}_2$ compares pooled statistics and so cannot tell
whether individual configurations are correct. For DW-4 and LJ-13, we
therefore also compare their whole configurations. Two configurations that
differ only by relabelling identical particles, or by a rotation or
reflection, describe the same physical state, so we measure the distance
between them after aligning out these symmetries. For configurations
$X, Y \in \mathbb{R}^{k\times s}$ ($k$ particles in $s$ dimensions,
centred at the origin), the squared distance is
\[
d^2(X,Y) = \min_{P,\,R}\; \|X - PYR\|_{\mathrm{F}}^2,
\]
where $P$ ranges over particle permutations and $R$ over rotations and
reflections. Geometric-$\mathcal{W}_2$ is the $2$-Wasserstein distance
between the generated and reference samples with $d$ as the ground
distance.

We approximate the joint alignment: following \citet{kohler2020equivariant}, we match particles by a linear assignment and then fit a rotation or reflection to the matched coordinates by least squares (orthogonal Procrustes). The resulting squared distance is an upper bound on $d^2(X,Y)$. We use these approximate alignment costs for the full sets of $10{,}000$ generated and $10{,}000$ reference samples. The outer transport problem is solved exactly for this cost matrix; the configuration alignment remains approximate.

\subsubsection{{Molecular system}}
{For Alanine Dipeptide we use the $10{,}000$-frame molecular-dynamics (MD) test split as the reference. SSS produces $10{,}000$ diffusion endpoints and $4{,}096$ buffer draws. We discard configurations with non-finite energy before evaluation. Both energy discrepancies use reduced energies $e(x)=U(x)/(k_BT_{\rm phys})$ and exact one-dimensional empirical transport, including when sample sizes differ. Energy-$\mathcal W_2$ is defined above.}

\paragraph{{Energy-$\mathcal W_1$.}}
{For generated configurations $x_1,\ldots,x_n$ and reference configurations $y_1,\ldots,y_m$, we compute
\[
 \text{Energy-}\mathcal W_1 =
 \mathcal W_1\!\left(\frac1n\sum_{i=1}^n\delta_{e(x_i)},\,
                      \frac1m\sum_{j=1}^m\delta_{e(y_j)}\right).
\]
The ground distance is the absolute difference between two reduced energies. The metric measures the average displacement needed to match the two energy distributions.}

\paragraph{{$\mathbb T$-$\mathcal W_2$.}}
{We represent each configuration by its backbone angles $a=(\phi,\psi)\in[-\pi,\pi)^2$. For angle pairs $a$ and $b$, we use the squared ground distance
\[
 d_{\mathbb T}^2(a,b)=\sum_{j=1}^{2}\min\{|a_j-b_j|,\,2\pi-|a_j-b_j|\}^2.
\]
We compute the $2$-Wasserstein distance between the empirical distributions of generated and reference angle pairs using this ground distance and uniform sample weights. Measuring each difference around the circle keeps angles just across the periodic boundary close together.}

\paragraph{{Other metrics in the PITA benchmark.}}
{We omit Rama-KL because its histogram estimate depends strongly on sample size. Even between MD samples, our estimated floor is about $1.5$ at $10{,}000$ samples and $3.0$ at $4{,}096$, the two sizes used here. The published benchmark does not specify the bins, smoothing constant or sample count needed to match its estimate. We also omit TICA-$\mathcal W_1$ and TICA-$\mathcal W_2$. Their projection is learned from pairs of MD frames 100 steps apart. The released 300 K splits are shuffled and have near-zero frame correlations at lags 1--100, so they do not preserve the information needed to reproduce that projection.}

\subsection{Numerical settings and baselines reimplementation}
\label{app:hyperparameters}
We list here the settings needed to reproduce our results. {The numerical diffusion update uses the step size of each grid interval.} The SMC particle count $N_{\mathrm{SMC}}$ also sets the replay-buffer capacity $C$, and {$K+1$ counts all temperatures in the adaptive sweep, including $\beta_0$} (Appendix~\ref{app:smc}). The fixed training lengths below were selected using validation performance, as were the baseline training lengths. We report the cost of these final checkpoints, rather than time to a shared error tolerance or a formal convergence threshold.

\paragraph{GMM16 ($d=16$).}
{The diffusion horizon is $T=4$.} The SMC sweep uses $N_{\mathrm{SMC}}=81{,}920$ particles, ESS threshold $\eta=0.5$, and $3$ HMC steps per temperature, {giving $K=11,12,12$ for seeds 0, 1, and 2, respectively}. Training runs for $10$ outer iterations. Each iteration generates $N=30{,}000$ proposals per temperature with $L_{\mathrm{train}}=100$ sampler steps, applies $n_{\mathrm{MCMC}}=10$ HMC steps after resampling, and then takes $J=10^3$ Adam steps. For final evaluation, we generate $10^5$ fresh samples with $L_{\mathrm{eval}}=2000$ steps and compare them with $10^5$ exact draws from the target.
\paragraph{GMM50 ($d=50$).}
{The diffusion horizon is $T=4$.} The per-coordinate target is the same as in GMM16, but the higher dimension calls for more MCMC effort at initialization and longer training. The SMC sweep uses $N_{\mathrm{SMC}}=81{,}920$, $\eta=0.1$, and $50$ HMC steps per temperature, giving $K=11$ for all three seeds. Training runs for $10$ outer iterations with $N=30{,}000$, $L_{\mathrm{train}}=100$, $n_{\mathrm{MCMC}}=20$, and $J=10^4$. Evaluation follows GMM16, with $10^5$ fresh samples, $L_{\mathrm{eval}}=2000$, and $10^5$ exact reference draws.
\paragraph{DW-4 ($d=8$).}
{The diffusion horizon is $T=2$.} The SMC sweep uses $N_{\mathrm{SMC}}=81{,}920$, $\eta=0.2$, and $10$ HMC steps per temperature, giving $K=6$ for all three seeds. Training runs for $10$ outer iterations with $N=30{,}000$, $L_{\mathrm{train}}=100$, $n_{\mathrm{MCMC}}=10$, and $J=10^3$. To keep training stable under large energy gradients, we cap the norm of the regression target at $20$ and clamp the network output to $[-10^4,10^4]$. For final evaluation, we generate $10^4$ fresh samples with $L_{\mathrm{eval}}=2000$ steps.
\paragraph{LJ-13 ($d=39$).}
{The diffusion horizon is $T=1.5$.} The SMC sweep uses $N_{\mathrm{SMC}}=25{,}000$, $\eta=0.2$, and $200$ HMC steps per temperature, giving $K=7$ for all three seeds. Training runs for $50$ outer iterations with $N=5{,}000$, {$L_{\mathrm{train}}=55$}, $n_{\mathrm{MCMC}}=2000$, and $J=10^3$. We cap the norm of the regression target at $150$ per particle and clamp the network output to $[-10^4,10^4]$. For final evaluation, we generate $10^4$ fresh samples with {$L_{\mathrm{eval}}=2000$ steps.}

\paragraph{{Alanine Dipeptide.}}
{The diffusion horizon is $T=2$.} {SMC uses $4{,}096$ particles, an ESS threshold of $0.8$, and 50 periodic HMC moves per level in internal coordinates. All three seeds give 40 levels including the reference. We recalibrate each step size on its final SMC population for use during training. Each of 100 outer iterations generates $1{,}024$ paths per active level with 100 diffusion steps, applies 50 HMC moves, and takes 500 optimizer steps using 64 samples per buffer. We write a new block only if it has a finite path weight and the path-weight ESS is at least $1\%$ of the batch size. Final evaluation uses 1,000 diffusion steps.}

\paragraph{{Alanine Dipeptide training loss.}}\label{app:aldp-loss}
{We draw $t$ log-uniformly on $[t_{\min},T]=[10^{-5},2]$. For the prediction $r=r_\theta(X_t,t,\beta_k)$ and the two regression targets in Section~\ref{sec:regression}, we minimize
\[
\E\!\left[\sigma_t^2\left(\norm{r-\varphi_{\rm DSM}}^2+
0.01\,\mathbf 1_{\{t<t_{\mathrm{threshold}}\}}\norm{r-\varphi_{\rm TSM}}^2\right)\right],
\qquad t_{\mathrm{threshold}}=0.0184.
\]
We project the noise onto the zero center-of-mass subspace and clip the clean relative score in the TSM target as described in Appendix~\ref{app:architecture}.}

\paragraph{{Temperature-based baselines.}}
{We reimplement \PITA{} and \TABG{} and report our runs on GMM16, GMM50, and DW-4. On LJ-13 and Alanine Dipeptide, both methods diverged on our hardware. For those two targets we use their published discrepancies from \citet{akhoundsadegh2025progressive}; local timing and evaluation budgets are unavailable for those entries. On the three successful targets, both methods use the positive inverse temperatures selected by the SSS warm-up for each seed. On both mixtures, each of PITA's two networks uses the same residual MLP as SSS. On DW-4, each uses PITA's EGNN with its width adjusted to within $0.8\%$ of the SSS parameter count. PITA therefore has about twice as many parameters in total. TA-BG keeps its spline flow, with conditioner widths chosen so that the entire flow matches SSS within $0.9\%$.

Both baselines use the same preconditioned HMC kernel as SSS. We allocate the same nominal number of particle proposals as SSS uses in its warm-up and training rejuvenation combined, distributed across annealing transitions. PITA applies HMC after each transition. TA-BG applies it after importance resampling and before fitting the flow at the next temperature.}

\paragraph{{Adjoint Sampling.}}
{We reimplement AS \citep{havens2025adjoint} on all five targets and use an HMC-augmented version for the reported comparisons. For each seed, the final-temperature population from SSS's SMC initialization seeds a uniformly weighted FIFO buffer. AS trains only at $\beta=1$. Before each new batch enters the buffer, we apply the HMC kernel tuned by SMC at that temperature, with the same number of moves per particle as in SSS training. AS uses neither path-space importance resampling nor temperature-bridge weights.

We use SSS's OU process in place of AS's variance-exploding process, so the endpoint regression target coincides with the SSS target-score label at $\beta=1$. For each target and seed, we match the network, time grid, optimizer, total diffusion paths, optimizer steps and total minibatch size. The AS buffer has the capacity of one SSS buffer. Including initialization, AS used $0.005$--$2\%$ fewer energy-gradient evaluations than SSS.}

\subsection{{Computational cost}}\label{app:costs}
{Table~\ref{tab:evals} gives evaluation counts for the benchmark comparisons. Table~\ref{tab:time} reports the available wall-clock measurements. Counts vary across seeds because the number of temperatures and the work needed to tune the kernel can vary. Thus, the reported runtimes and budget counts are averaged over 3 random seeds that we ran to produce the results. For PITA, TA-BG, and 
\method{}, they use the same temperature ladder, so we run the methods until there is no improvement on the metrics or training loss. For AS, it does not contain any temperature setting, so we match the energy-evaluation of the runs with those in \method{}. This will allow a much larger budget to be spent on AS compared to \method{} at the target temperature $\beta = 1$.}
\begin{table}[t]
\centering
\caption{{Wall-clock time (in minutes) for the separately timed runs. All runs were executed on a single NVIDIA RTX 4500 Ada Generation GPU. The LJ-13 and Alanine Dipeptide results of \TABG{} and \PITA{} are taken from \citet{akhoundsadegh2025progressive}, so no timing is reported.}}
\label{tab:time}
\small
\begin{tabular}{lrrrr}
\toprule
\textsc{Target} & \TABG{} & \PITA{} & SSS & AS \\
\midrule
GMM16 ($d=16$) & $40$ & $105$ & $6$ & $2$ \\
GMM50 ($d=50$) & $55$ & $190$ & $43$ & $6$ \\
DW-4 ($d=8$) & $80$ & $530$ & $15$ & $13$ \\
LJ-13 ($d=39$) & --- & --- & $552$ & $538$ \\
Alanine Dipeptide & --- & --- & $366$ & $351$ \\
\bottomrule
\end{tabular}
\end{table}
\begin{table}[htbp]
\centering\small
\caption{{Energy and energy-gradient evaluations, averaged over three seeds. AS used matched SSS evaluation budgets on all five targets. Local budgets for \TABG{} and \PITA{} are unavailable on LJ-13 and Alanine Dipeptide.}}
\label{tab:evals}
\setlength{\tabcolsep}{7pt}
\begin{lrbox}{\resultbox}\begin{tabular}{@{}lrrr@{}}
\toprule
Target & Method & Energy (millions) & Gradient (millions) \\
\midrule
GMM16 & SSS (ours) & 85.7 & 41.7 \\
 & PITA & 411.6 & 368.4 \\
 & TA-BG & 86.9 & 44.9 \\
 & AS & Matched SSS & Matched SSS \\
\midrule
GMM50 & SSS (ours) & 230.5 & 114.1 \\
 & PITA & 551.2 & 438.6 \\
 & TA-BG & 227.0 & 115.3 \\
 & AS & Matched SSS & Matched SSS \\
\midrule
DW-4 & SSS (ours) & 50.2 & 24.5 \\
 & PITA & 374.6 & 350.7 \\
 & TA-BG & 98.0 & 73.8 \\
 & AS & Matched SSS & Matched SSS \\
\midrule
LJ-13 & SSS (ours) & 9,077 & 3,038 \\
 & AS & Matched SSS & Matched SSS \\
\midrule
Alanine Dipeptide & SSS (ours) & 420.2 & 208.1 \\
 & AS & Matched SSS & Matched SSS \\
\bottomrule\end{tabular}\end{lrbox}
\ifdim\wd\resultbox>\linewidth\resizebox{\linewidth}{!}{\usebox{\resultbox}}\else\usebox{\resultbox}\fi
\end{table}
\subsection{{Discussion on practical implementation}}\label{app:engineering}
\paragraph{{Initialization.}}
{One adaptive SMC pass selects the temperature ladder, seeds the buffers, and tunes a local kernel at each positive temperature. Training reuses these choices; for Alanine Dipeptide we recalibrate the step sizes on the final SMC populations. A stronger initialization costs more target evaluations up front, but can preserve modes that later local moves rarely reach. Initial mode weights need not be exact. In the selected eccentric-mixture run in Figure~\ref{fig:far-retention}, SSS brought an underrepresented mode from $0.23$ to $0.49$; when the smaller initial population lost that mode, training did not recover it. This distinction helps explain why increasing the particle count matters at a bottleneck and has little effect on corrected samples once DW-4 is adequately covered.}

\paragraph{{Importance-weight variance.}}
{The SMC ladder controls the change between adjacent targets through a bridge ESS criterion. Path weights depend on the learned proposal and can still concentrate even when adjacent targets overlap well. Early in training, the buffers retain SMC particles while fresh batches gradually replace them. At each refresh we resample the current paths, rejuvenate the endpoints, and normalize the new bridge weights within the batch, so path weights are never multiplied across outer iterations. Within a path, we use finer time steps near the endpoint, where the low-noise updates can contribute substantial weight variability. The grid is exponential on the mixtures, DW-4, and Alanine Dipeptide, and quadratic on LJ-13. These choices still leave few distinct endpoints when path ESS is low. On LJ-13 and Alanine Dipeptide, the final buffer gives lower energy discrepancies than resampling fresh paths alone.}

\paragraph{{Rejuvenation length.}}
{When resampling produces many copies of a few endpoints, longer chains give those copies more opportunity to explore their local basin. Each additional move costs target evaluations. In a separate LJ-13 chain-length experiment, increasing rejuvenation from 20 to 100 and then 2,000 moves reduced buffer Energy-$\mathcal W_2$ from $2.31$ to $0.37$ and then $0.27$. For fresh importance-corrected samples, the corresponding errors were $0.74$, $0.40$, and $0.36$. Most of the observed reduction came in the first increase. Longer local chains can be worthwhile when duplicated particles are the main difficulty; when a whole mode is missing, allocating more particles to initialization can be more effective.}

\section{{Additional baselines}}
In this Section, we consider two additional baselines for \method{}: Initial sampler and Long run SMC.

\subsection{{Initial sampler}}\label{app:initial-sampler}
{Before the first training update, the final buffer $\Data_K$ contains the $\beta=1$ population from the adaptive SMC sweep. We evaluate this population on the same three seeds and against the same reference samples as the trained SSS outputs in Table~\ref{tab:reported}. Table~\ref{tab:initial-sampler} compares the four sample sets.}

\begin{table}[htbp]
\centering
\footnotesize
\caption{{Initial SMC population and SSS samples after training. Entries are mean $\pm$ standard deviation over three seeds, rounded to two decimal places. For each target and metric, \textbf{bold} marks the lowest displayed mean and \underline{underlining} marks the second-lowest distinct mean. Equal displayed means share a rank. A dash denotes an inapplicable metric and $\ast$ denotes a mean above $10^4$.}}\label{tab:initial-sampler}
\setlength{\tabcolsep}{3pt}
\begin{lrbox}{\resultbox}\begin{tabular}{@{}llrrr@{}}
\toprule
Target & Samples & Distance-$\mathcal W_2$ & Energy-$\mathcal W_2$ & Geometric-$\mathcal W_2$ \\
\midrule
GMM16 ($d=16$) & SMC (initial) & $0.37 \pm 0.15$ & $0.61 \pm 0.41$ & \NA{} \\
 & SSS (raw) & $0.13 \pm 0.09$ & $1.15 \pm 0.12$ & \NA{} \\
 & SSS (corrected) & $\mathbf{0.01} \pm 0.01$ & $\mathbf{0.04} \pm 0.00$ & \NA{} \\
 & SSS (buffer) & $\underline{0.04} \pm 0.01$ & $\underline{0.05} \pm 0.02$ & \NA{} \\
\midrule
GMM50 ($d=50$) & SMC (initial) & $0.57 \pm 0.12$ & $3.93 \pm 1.23$ & \NA{} \\
 & SSS (raw) & $\underline{0.17} \pm 0.03$ & $4.30 \pm 0.81$ & \NA{} \\
 & SSS (corrected) & $\mathbf{0.05} \pm 0.02$ & $\mathbf{0.32} \pm 0.06$ & \NA{} \\
 & SSS (buffer) & $0.21 \pm 0.04$ & $\underline{1.03} \pm 0.20$ & \NA{} \\
\midrule
DW-4 ($d=8$) & SMC (initial) & $\mathbf{0.11} \pm 0.12$ & $\underline{0.17} \pm 0.02$ & $\mathbf{0.38} \pm 0.19$ \\
 & SSS (raw) & $0.23 \pm 0.07$ & $0.92 \pm 0.31$ & $0.52 \pm 0.14$ \\
 & SSS (corrected) & $\underline{0.19} \pm 0.01$ & $\mathbf{0.11} \pm 0.03$ & $\underline{0.48} \pm 0.01$ \\
 & SSS (buffer) & $\underline{0.19} \pm 0.01$ & $\mathbf{0.11} \pm 0.01$ & $\underline{0.48} \pm 0.02$ \\
\midrule
LJ-13 ($d=39$) & SMC (initial) & $\mathbf{0.00} \pm 0.00$ & $\mathbf{0.23} \pm 0.06$ & $\mathbf{1.46} \pm 0.00$ \\
 & SSS (raw) & $0.04 \pm 0.03$ & $\ast$ & $1.50 \pm 0.04$ \\
 & SSS (corrected) & $0.03 \pm 0.03$ & $1.41 \pm 1.73$ & $1.63 \pm 0.06$ \\
 & SSS (buffer) & $\underline{0.01} \pm 0.00$ & $\underline{0.32} \pm 0.06$ & $\underline{1.47} \pm 0.00$ \\
\bottomrule\end{tabular}\end{lrbox}
\ifdim\wd\resultbox>\linewidth\resizebox{\linewidth}{!}{\usebox{\resultbox}}\else\usebox{\resultbox}\fi
\end{table}

{On both Gaussian mixtures, corrected SSS samples have smaller mean distance and energy errors than the initialization. For example, on GMM50 the distance error falls from $0.57$ to $0.05$ and the energy error from $3.93$ to $0.32$. On DW-4, correction and buffering reduce the mean energy error, while the initial SMC population has lower mean distance and geometric errors. On LJ-13, initialization already gives the lowest mean discrepancies; the final buffer comes close, but training does not improve accuracy at $\beta=1$. The trained model still provides a reusable way to generate fresh proposals at any temperature on the ladder.}

\subsection{{Matched Energy budget SMC}}\label{app:smc-compare}
{We give a single SMC run the particle count, tempering rule, and preconditioned HMC kernel of the SSS warm-up. We increase the number of moves per temperature ($m$) to match the energy-evaluation budget of the entire SSS run, including initialization and training. This asks how much accuracy the same energy budget buys when it is spent entirely on SMC.}
\begin{table}[htbp]
\centering
\footnotesize
\caption{{Final SMC samples and SSS buffer samples (mean $\pm$ standard deviation over three seeds). Bold marks the lower mean for each target and metric.}}
\label{tab:smc-buffer}
\setlength{\tabcolsep}{3pt}
\begin{lrbox}{\resultbox}\begin{tabular}{@{}lrrrr@{}}
\toprule
Target & Samples & Distance-$\mathcal W_2$ & Energy-$\mathcal W_2$ & Geometric-$\mathcal W_2$ \\
\midrule
GMM16 & SMC ($m=40$) & $0.141 \pm 0.038$ & $0.188 \pm 0.062$ & \NA{} \\
 & SSS (buffer) & $\mathbf{0.041} \pm 0.009$ & $\mathbf{0.048} \pm 0.016$ & \NA{} \\
\midrule
GMM50 & SMC ($m=120$) & $0.383 \pm 0.128$ & $2.240 \pm 1.020$ & \NA{} \\
 & SSS (buffer) & $\mathbf{0.208} \pm 0.037$ & $\mathbf{1.025} \pm 0.201$ & \NA{} \\
\midrule
DW-4 & SMC ($m=45$) & $\mathbf{0.081} \pm 0.033$ & $0.124 \pm 0.024$ & $\mathbf{0.312} \pm 0.041$ \\
 & SSS (buffer) & $0.194 \pm 0.010$ & $\mathbf{0.105} \pm 0.014$ & $0.479 \pm 0.016$ \\
\midrule
LJ-13 & SMC ($m=25{,}927$) & $\mathbf{0.005} \pm 0.001$ & $\mathbf{0.262} \pm 0.044$ & $\mathbf{1.461} \pm 0.001$ \\
 & SSS (buffer) & $0.006 \pm 0.002$ & $0.324 \pm 0.063$ & $1.465 \pm 0.003$ \\
\bottomrule\end{tabular}\end{lrbox}
\ifdim\wd\resultbox>\linewidth\resizebox{\linewidth}{!}{\usebox{\resultbox}}\else\usebox{\resultbox}\fi
\end{table}
{On GMM16, the buffer reduces the mean distance error from $0.141$ to $0.041$ and the energy error from $0.188$ to $0.048$. Both errors are roughly halved on GMM50 (Table~\ref{tab:smc-buffer}). Once the mixture modes separate, extra local moves can explore each occupied mode while doing little to adjust how many particles represent it. Fresh proposals and importance correction give SSS another opportunity to change those proportions. The particle systems give a different result. On DW-4, SMC has lower distance and geometric errors, while the buffer has a slightly lower energy error. On LJ-13, SMC has lower means on all three metrics, with geometric errors close at $1.461$ and $1.465$.}

\paragraph{{Amortizing fixed SMC samples.}}
{On the Gaussian mixtures, we also train the same temperature-conditioned diffusion on fixed SMC samples at each temperature. We keep the target score-matching loss, optimizer, and training schedule of SSS, and turn off buffer refreshes. Storing the energy gradients adds at most about $1\%$ to the SMC energy-evaluation count. We compare fresh raw samples from both models. On GMM50, refreshing the data during training reduces mean distance error from $0.398$ to $0.166$. On GMM16, the difference is smaller and the energy errors are similar.}
\begin{table}[htbp]
\centering\small
\caption{{Raw diffusion samples after training on fixed SMC populations or with SSS (mean $\pm$ standard deviation over three seeds). Bold marks the lower mean for each target and metric.}}
\label{tab:smc-amort}
\setlength{\tabcolsep}{3pt}
\begin{lrbox}{\resultbox}\begin{tabular}{@{}lrrr@{}}
\toprule
Target & Sampler & Distance-$\mathcal W_2$ & Energy-$\mathcal W_2$ \\
\midrule
GMM16 & SMC + amortisation & $0.188 \pm 0.042$ & $1.166 \pm 0.106$ \\
 & SSS (raw) & $\mathbf{0.130} \pm 0.091$ & $\mathbf{1.153} \pm 0.121$ \\
\midrule
GMM50 & SMC + amortisation & $0.398 \pm 0.125$ & $5.462 \pm 0.873$ \\
 & SSS (raw) & $\mathbf{0.166} \pm 0.034$ & $\mathbf{4.304} \pm 0.807$ \\
\bottomrule\end{tabular}\end{lrbox}
\ifdim\wd\resultbox>\linewidth\resizebox{\linewidth}{!}{\usebox{\resultbox}}\else\usebox{\resultbox}\fi
\end{table}
\section{{Additional ablation studies}}\label{app:additional}
{\textbf{Question 1 (Keeping a fixed buffer).} Does fixing $\Data_1$ help by preserving an accurate starting population, and how much does an error in that buffer affect the final samples?}

{On DW-4, the first positive inverse temperature is close to the reference ($\beta_1\approx0.036$). We compare keeping $\mathcal D_1$ fixed with refreshing it from corrected model samples or from fresh reference draws reweighted to $\beta_1$. We also fix it at deliberately inaccurate standard Gaussian draws. Even in the last case, where the first-buffer discrepancy rises to $0.667$, the final corrected-sample errors remain close to the default (Table~\ref{tab:anchor}). Refreshing from the model gives slightly larger mean errors on all three metrics. Drawing fresh reference samples gives a similar distance error to the fixed buffer. These results support keeping the first buffer fixed on DW-4, where the first rung is already close to the reference.}
\begin{table}[htbp]
\centering\small
\caption{{First-buffer ablation on DW-4. The first three metrics summarize corrected samples over three seeds. The final column measures the discrepancy between the first buffer and $\pi_{\beta_1}$.}}
\label{tab:anchor}
\setlength{\tabcolsep}{3pt}
\begin{lrbox}{\resultbox}\begin{tabular}{@{}lrrrr@{}}
\toprule
$\mathcal D_1$ & Distance-$\mathcal W_2$ & Energy-$\mathcal W_2$ & Geometric-$\mathcal W_2$ & $\mathcal W_2(\mathcal D_1)$ \\
\midrule
Fixed (ours) & $0.192 \pm 0.010$ & $0.109 \pm 0.034$ & $0.482 \pm 0.014$ & $0.008$ \\
Self-refresh & $0.208 \pm 0.015$ & $0.126 \pm 0.025$ & $0.507 \pm 0.024$ & $0.007$ \\
Reference bridge & $0.189 \pm 0.049$ & $0.138 \pm 0.014$ & $0.484 \pm 0.082$ & $0.008$ \\
Bad anchor & $0.196 \pm 0.023$ & $0.122 \pm 0.015$ & $0.486 \pm 0.037$ & $0.667$ \\
\bottomrule\end{tabular}\end{lrbox}
\ifdim\wd\resultbox>\linewidth\resizebox{\linewidth}{!}{\usebox{\resultbox}}\else\usebox{\resultbox}\fi
\end{table}

{\textbf{Question 2 (Choosing the number of SMC particles).} How sensitive is SSS to the particle count when initialization covers the important configurations? What changes when a mode becomes rare along the tempering path?}

{On DW-4, SMC already covers all the modes at the smallest particle count tested. Corrected SSS errors change little as $N$ increases from $1{,}000$ to $81{,}920$. For instance, the corrected geometric error stays between $0.48$ and $0.51$ across the sweep (Table~\ref{tab:particle-count}). This suggests that the benefit of extra particles depends on how well initialization covers the target. The SMC warm-up took 10--13 seconds at both endpoints on our GPU.}
\begin{table}[htbp]
\centering\small
\caption{{Effect of SMC particle count on DW-4. Entries are mean $\pm$ standard deviation.}}
\label{tab:particle-count}
\setlength{\tabcolsep}{3pt}
\begin{lrbox}{\resultbox}\begin{tabular}{@{}lrrrr@{}}
\toprule
Samples & $N$ & Distance-$\mathcal W_2$ & Energy-$\mathcal W_2$ & Geometric-$\mathcal W_2$ \\
\midrule
SSS (corrected) & 1,000 & $0.180 \pm 0.010$ & $0.100 \pm 0.030$ & $0.480 \pm 0.010$ \\
 & 3,000 & $0.200 \pm 0.010$ & $0.140 \pm 0.030$ & $0.490 \pm 0.010$ \\
 & 5,000 & $0.200 \pm 0.010$ & $0.120 \pm 0.010$ & $0.490 \pm 0.010$ \\
 & 10,000 & $0.210 \pm 0.020$ & $0.140 \pm 0.030$ & $0.510 \pm 0.030$ \\
 & 20,000 & $0.200 \pm 0.020$ & $0.120 \pm 0.020$ & $0.500 \pm 0.030$ \\
 & 50,000 & $0.190 \pm 0.010$ & $0.130 \pm 0.030$ & $0.490 \pm 0.020$ \\
 & 81,920 & $0.190 \pm 0.010$ & $0.110 \pm 0.030$ & $0.480 \pm 0.010$ \\
\midrule
SMC & 1,000 & $0.350 \pm 0.070$ & $0.450 \pm 0.250$ & $1.140 \pm 0.410$ \\
 & 3,000 & $0.330 \pm 0.180$ & $0.140 \pm 0.010$ & $0.910 \pm 0.330$ \\
 & 5,000 & $0.270 \pm 0.170$ & $0.130 \pm 0.030$ & $0.690 \pm 0.280$ \\
 & 10,000 & $0.350 \pm 0.160$ & $0.200 \pm 0.120$ & $0.850 \pm 0.310$ \\
 & 20,000 & $0.400 \pm 0.200$ & $0.090 \pm 0.050$ & $0.880 \pm 0.370$ \\
 & 50,000 & $0.070 \pm 0.030$ & $0.100 \pm 0.040$ & $0.330 \pm 0.070$ \\
 & 81,920 & $0.110 \pm 0.120$ & $0.180 \pm 0.020$ & $0.380 \pm 0.190$ \\
\bottomrule\end{tabular}\end{lrbox}
\ifdim\wd\resultbox>\linewidth\resizebox{\linewidth}{!}{\usebox{\resultbox}}\else\usebox{\resultbox}\fi
\end{table}

{To study the second question, we use the Eccentric GMM. In dimension $d=50$, its target density is
\[
\pi(x)=\dfrac12\,\mathcal N\!\left(x;0,\dfrac1{16}I_d\right)
      +\dfrac12\,\mathcal N\!\left(x;\dfrac34\mathbf 1_d,\dfrac1{16}I_d\right),
\qquad \rho=\mathcal N(0,I_d),
\]
where $\mathbf 1_d$ is the vector of ones. We call the component centered at $\tfrac34\mathbf 1_d$ the far mode because its center lies farther from the center of the reference distribution $\rho$. Quadrature gives a far-mode mass of only $1.2\times10^{-4}$ near $\beta=0.2$, although its mass at $\beta=1$ is $0.5$. We ran SMC over 100 seeds at each particle count, matching the energy budget of $80{,}000$ particles with ten HMC moves per temperature. Smaller populations received more HMC moves. Table~\ref{tab:bottleneck} shows that increasing the particle count from $1{,}000$ to $80{,}000$ raised the fraction of runs retaining the far mode from $3\%$ to $85\%$. At this bottleneck, longer local chains did little to compensate for a small population.}
\begin{table}[htbp]
\centering\small
\caption{{Far-mode retention in the eccentric Gaussian mixture at a fixed SMC energy budget. Each row summarizes 100 runs.}}\label{tab:bottleneck}
\begin{tabular}{rrr}\toprule
Particles & HMC moves per temperature & Runs retaining the far mode \\
\midrule
1,000 & 1,000 & 3\% \\
3,000 & 330 & 13\% \\
5,000 & 195 & 25\% \\
10,000 & 97 & 36\% \\
20,000 & 47 & 50\% \\
50,000 & 17 & 71\% \\
80,000 & 10 & 85\% \\
\bottomrule\end{tabular}
\end{table}

\begin{figure}[htbp]
\centering
\includegraphics[width=.78\linewidth]{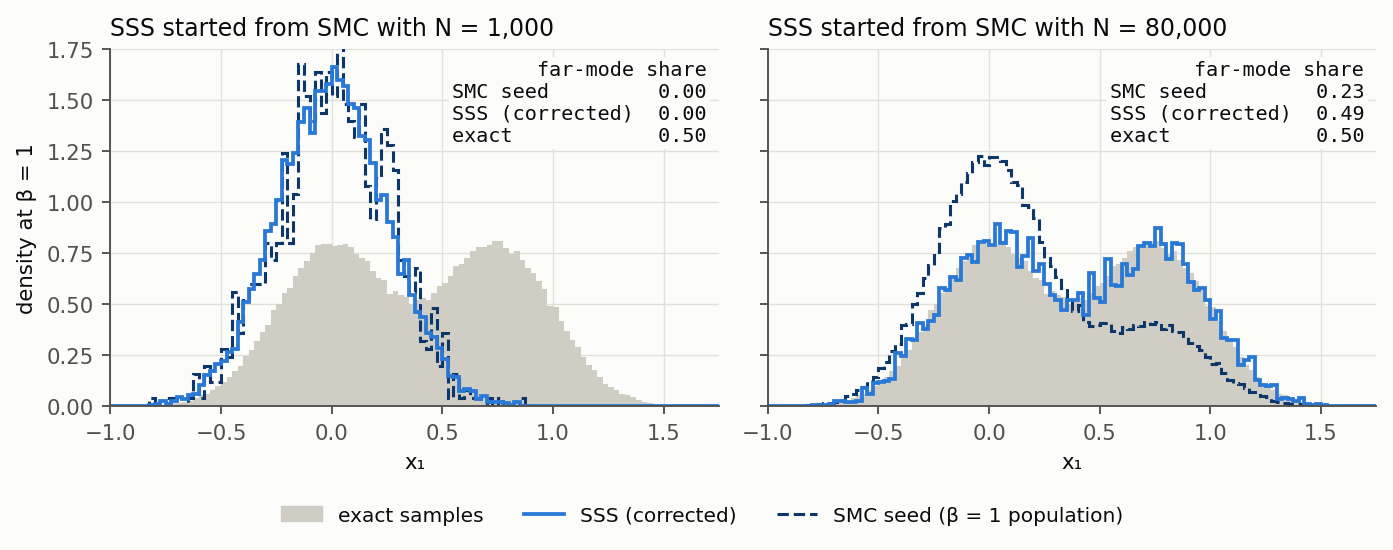}
\caption{{Marginal densities of $x_1$ at $\beta=1$ for the selected seed, starting from SMC populations of $1{,}000$ (left) and $80{,}000$ (right) particles. Each panel compares the initial SMC population, corrected SSS samples, and exact samples. Insets report the corresponding far-mode shares, with exact target mass $0.50$.}}\label{fig:far-retention}
\end{figure}

{We then followed SSS training for one seed chosen to illustrate the two outcomes (Figure~\ref{fig:far-retention}). With $80{,}000$ particles, SMC retained the far mode but assigned it mass $0.23$. Within a few outer iterations SSS brought that share to $0.49$, close to the exact $0.50$. With $1{,}000$ particles, SMC lost the mode and SSS cannot recover it.}

{\textbf{Question 3 (Choosing the MCMC kernel).} How much does the choice of MCMC kernel affect SSS?}

{We replace the kernel in both the SMC warm-up and the training rejuvenation steps. Each kernel uses the same empirical covariance preconditioner and number of moves per temperature. We tune the step size by bisection, targeting acceptance rates of $0.8$--$0.95$ for one-step HMC with partial momentum refresh, $0.4$--$0.65$ for MALA, and $0.2$--$0.35$ for random-walk Metropolis--Hastings (RWMH).

Kernel choice has relatively little effect on GMM16 and DW-4. On GMM50, the two gradient-based kernels give much smaller energy errors than RWMH. On LJ-13, HMC gives the lowest corrected-sample errors on all three metrics, while buffer geometric errors are similar across kernels (Tables~\ref{tab:kernel-corrected} and~\ref{tab:kernel-buffer}). The gap is larger for fresh corrected proposals than for the stored LJ-13 population. This distinction matters when the goal is to train a sampler for repeated use.}
\begin{table}[h]
\centering
\footnotesize
\caption{{Kernel comparison for SSS corrected samples (mean $\pm$ standard deviation over three seeds). For each target and metric, bold marks the lowest displayed mean and underlining marks the second-lowest distinct mean.}}
\label{tab:kernel-corrected}
\setlength{\tabcolsep}{3pt}
\begin{lrbox}{\resultbox}\begin{tabular}{@{}lrrrr@{}}
\toprule
Target & Kernel & Distance-$\mathcal W_2$ & Energy-$\mathcal W_2$ & Geometric-$\mathcal W_2$ \\
\midrule
GMM16 & HMC (default) & $\mathbf{0.005} \pm 0.005$ & $0.040 \pm 0.004$ & \NA{} \\
 & MALA & $0.017 \pm 0.005$ & $\mathbf{0.031} \pm 0.005$ & \NA{} \\
 & RWMH & $\underline{0.016} \pm 0.014$ & $\underline{0.033} \pm 0.007$ & \NA{} \\
\midrule
DW-4 & HMC (default) & $\underline{0.192} \pm 0.010$ & $\mathbf{0.109} \pm 0.034$ & $\underline{0.482} \pm 0.014$ \\
 & MALA & $\mathbf{0.191} \pm 0.004$ & $0.137 \pm 0.006$ & $\mathbf{0.477} \pm 0.008$ \\
 & RWMH & $0.200 \pm 0.012$ & $\underline{0.118} \pm 0.007$ & $0.495 \pm 0.023$ \\
\midrule
GMM50 & HMC (default) & $\underline{0.048} \pm 0.022$ & $\mathbf{0.321} \pm 0.059$ & \NA{} \\
 & MALA & $\mathbf{0.033} \pm 0.023$ & $\underline{0.369} \pm 0.199$ & \NA{} \\
 & RWMH & $0.360 \pm 0.071$ & $7.731 \pm 5.996$ & \NA{} \\
\midrule
LJ-13 & HMC (default) & $\mathbf{0.026} \pm 0.027$ & $\mathbf{1.405} \pm 1.733$ & $\mathbf{1.626} \pm 0.059$ \\
 & MALA & $0.200 \pm 0.283$ & $14.460 \pm 18.850$ & $2.205 \pm 0.659$ \\
 & RWMH & $\underline{0.125} \pm 0.189$ & $\underline{8.320} \pm 12.770$ & $\underline{2.102} \pm 0.758$ \\
\bottomrule\end{tabular}\end{lrbox}
\ifdim\wd\resultbox>\linewidth\resizebox{\linewidth}{!}{\usebox{\resultbox}}\else\usebox{\resultbox}\fi
\end{table}
\begin{table}[h]
\centering
\footnotesize
\caption{{Kernel comparison for SSS buffer samples (mean $\pm$ standard deviation over three seeds). For each target and metric, bold marks the lowest displayed mean and underlining marks the second-lowest distinct mean. Equal displayed means share a rank.}}
\label{tab:kernel-buffer}
\setlength{\tabcolsep}{3pt}
\begin{lrbox}{\resultbox}\begin{tabular}{@{}lrrrr@{}}
\toprule
Target & Kernel & Distance-$\mathcal W_2$ & Energy-$\mathcal W_2$ & Geometric-$\mathcal W_2$ \\
\midrule
GMM16 & HMC (default) & $0.041 \pm 0.009$ & $0.048 \pm 0.016$ & \NA{} \\
 & MALA & $\mathbf{0.019} \pm 0.013$ & $\mathbf{0.042} \pm 0.005$ & \NA{} \\
 & RWMH & $\underline{0.026} \pm 0.028$ & $\underline{0.044} \pm 0.014$ & \NA{} \\
\midrule
DW-4 & HMC (default) & $\mathbf{0.194} \pm 0.010$ & $\mathbf{0.105} \pm 0.014$ & $\mathbf{0.479} \pm 0.016$ \\
 & MALA & $\underline{0.200} \pm 0.006$ & $\underline{0.114} \pm 0.024$ & $\underline{0.489} \pm 0.007$ \\
 & RWMH & $0.204 \pm 0.013$ & $0.135 \pm 0.007$ & $0.490 \pm 0.020$ \\
\midrule
GMM50 & HMC (default) & $\underline{0.208} \pm 0.037$ & $\underline{1.025} \pm 0.201$ & \NA{} \\
 & MALA & $\mathbf{0.136} \pm 0.124$ & $\mathbf{0.589} \pm 0.650$ & \NA{} \\
 & RWMH & $0.430 \pm 0.322$ & $6.069 \pm 1.520$ & \NA{} \\
\midrule
LJ-13 & HMC (default) & $\mathbf{0.006} \pm 0.002$ & $\mathbf{0.324} \pm 0.063$ & $\underline{1.465} \pm 0.003$ \\
 & MALA & $\underline{0.012} \pm 0.010$ & $0.501 \pm 0.319$ & $1.472 \pm 0.010$ \\
 & RWMH & $\underline{0.012} \pm 0.001$ & $\underline{0.409} \pm 0.160$ & $\mathbf{1.463} \pm 0.005$ \\
\bottomrule\end{tabular}\end{lrbox}
\ifdim\wd\resultbox>\linewidth\resizebox{\linewidth}{!}{\usebox{\resultbox}}\else\usebox{\resultbox}\fi
\end{table}
\end{document}